\documentclass[sn-mathphys-ay]{sn-jnl}

\usepackage{graphicx}
\usepackage{amsmath,amssymb,latexsym,bm}
\usepackage{subcaption}
\usepackage{cleveref}
\usepackage{amsthm}
\usepackage{mathrsfs}
\usepackage[title]{appendix}
\usepackage{multirow}
\usepackage{epsfig}
\usepackage{algorithmicx}
\usepackage{algorithm}
\usepackage{algpseudocode}

\DeclareMathOperator*{\maximize}{maximize}
\DeclareMathOperator*{\median}{median}

\newtheorem{theorem}{Theorem}
\newtheorem{corollary}{Corollary}

\begin{document}

	\title{An Adaptive Importance Sampling for Locally Stable Point Processes}

	
	\author[1]{\fnm{Hee-Geon} \sur{Kang}}\email{hgkang28@uos.ac.kr}
	\author*[1]{\fnm{Sunggon} \sur{Kim}}\email{sgkim@uos.ac.kr}
	
	\affil[1]{\orgdiv{ Department of Statistics}, \orgname{University of Seoul}, \orgaddress{\street{163 Seoulsiripdaero}, \city{Dongdaemun-gu}, \postcode{02504}, \state{Seoul}, \country{Republic of Korea}}}

	\abstract{
		The problem of finding the expected value of a statistic of a locally stable point process in a bounded region is addressed.  We propose an adaptive importance sampling for solving the problem. In our proposal, we restrict the  importance point process to the family of homogeneous Poisson point processes, which enables us to generate quickly independent samples of the importance point process.  The optimal intensity of the importance point process is found by applying the cross-entropy minimization method.  In the proposed scheme,  the expected value of the statistic and the optimal
		intensity are iteratively estimated in an adaptive manner.  We show that  the proposed estimator converges to the target value almost surely, and prove the asymptotic normality of it. We explain how to apply  the proposed scheme to the estimation of the intensity of a stationary pairwise interaction point process. The performance of the proposed scheme is compared numerically with Markov chain Monte Carlo simulation and perfect sampling.}

	
	\maketitle

	\section*{Statements and Declarations}
	{\bf Conflicts of interest} All authors declare that they have no conflicts of interest.

	\section{Introduction}
	Stationary pairwise interaction point processes are important stochastic models  in the study of spatial point patterns (Baddeley {\it et al.} \citeyearpar{baddeley2015spatial}). The summary characteristics such as the intensity and higher moments are usually unknown.  Such statistic can be estimated by restricting the process to a sufficiently large bounded region (Baddeley {\it et al.} \citeyearpar{baddeley2015spatial}, Illian  {\it et al.} \citeyearpar{illian2008statistical}). The probability density function (pdf) of a finite pairwise interaction point process is usually known up to constant. In order to find the explicit form of the normalization constant, the intensity, or some higher moments, we need to integrate a function of the point process with respect to the unnormalized pdf, which is not an easy task. The same is the case with locally stable point processes which are general point processes including stationary interaction point processes. The pdf of a locally stable point process in a bounded region  is also usually known up to constant, and it is  hard to find the explicit form of the normalization constant and the expected value of a statistic of the locally stable point process. Instead of finding exactly the integral value, one can resort to Monte Carlo methods such as Markov chain Monte Carlo and perfect sampling (M{\o}ller and Waagepetersen \citeyearpar{moller2003statistical}).

	In this paper,  we propose an adaptive importance sampling for finding the expected value of a statistic of a locally stable point process or a real-valued function of a locally stable point process in a bounded region. By restricting  the  importance sampling point process to the family of homogeneous Poisson point processes in our proposal, we can quickly generate  sample point processes.  The optimal intensity of the importance point process is found by applying the cross-entropy minimization method.  In the proposed scheme,  the expected value of the function and the optimal intensity are iteratively estimated in an adaptive manner.  We show that  the proposed estimator converges to the target value almost surely, and prove the asymptotic normality of it.
	
	Anderssen {\it et al.} \citeyearpar{anderssen2014numerical}  and Baddeley and Nair \citeyearpar{baddeley2012fast} proposed  the Poisson-saddlepoint approximation method to approximate the intensity of a stationary Gibbs point process. The method was extended to the case of a stationary Gibbs point process with any order by Baddeley and Nair \citeyearpar{baddeley2017poisson}.  In order to obtain sample point processes following the pdf of a locally stable point process, Markov chain Monte Carlo (MCMC) methods were proposed.  Geyer and M{\o}ller \citeyearpar{geyer1994simulation} and  M{\o}ller \citeyearpar{moller2019markov} proposed the birth-death Metropolis-Hastings algorithm.  Kelly and Ripley \citeyearpar{kelly1976note} and Ripley \citeyearpar{ripley1977modelling} proposed a spatial birth and death process. However, these methods may suffer from the slow convergence to the stationary distribution and a strong autocorrelation existing in generated sample point processes, which leads to a large number of samples required to obtain an accurate estimate to the target value. In order to obtain a random samples from the target distribution, perfect sampling is proposed (Robert et al.~\citeyearpar{robert1999monte}). It has been studied to apply perfect sampling to generate independent samples following exactly the pdf of a locally stable point process. To this end, Kendall and M{\o}ller \citeyearpar{kendall1999perfect,kendall2000perfect} gave a general formulation of the method of dominated coupling from the past (CFTP) and applied it to the problem of perfect sampling of locally stable point processes.  However, it takes a very long time to generate samples of point processes by applying perfect sampling  (M{\o}ller and Waagepetersen \citeyearpar{moller2003statistical}).
	
	Importance sampling  is a Monte Carlo method to estimate the integral value of a function with respect to the nominal distribution (Robert and Casella \citeyearpar{robert1999monte}, Rubinstein and Kroese \citeyearpar{rubinstein2016simulation}).  Instead of generating samples following the nominal distribution, it generates samples from an appropriately selected distribution called an importance sampling distribution.  By restricting the importance sampling distribution to a parametric family of distributions, one can find the optimal parameter of the importance sampling distribution.  Variance minimization and cross-entropy minimization methods are used generally to find the optimal parameter (De Boer {\it et al.} \citeyearpar{de2005tutorial}). However, in many situations, it is difficult to  find the optimal parameter explicitly, or it needs a large number of samples to estimate it accurately. Starting from a somewhat arbitrary  parameter, adaptive importance sampling adjusts the parameter of the importance sampling distribution during the sampling process (Bugallo  {\it et al.} \citeyearpar{bugallo2017adaptive}). In adaptive importance sampling, the importance sampling distribution is updated iteratively, and generates more samples of higher importance. In this way,  the efficiency of the estimation continues to improve with each iteration.  Oh and Berger \citeyearpar{oh1992adaptive} showed the convergence of the adaptive importance sampling estimator to the target value and the asymptotic normality of it.  We extend their results to the case of locally stable point processes defined in a bounded region.

	This paper is organized as follows. In Section~\ref{sec:MC methods}, we describe the birth-death Metropolis-Hastings algorithm and the dominated CFTP algorithm to estimate the expected value of a function of a locally stable point process in a bounded region.  We introduce our proposed estimator, and prove some asymptotic properties of it in Sections~\ref{sec:the proposed estimator} and~\ref{sec:asymptotic properties  of the proposed estimators}. We explain how to apply the proposed scheme to estimate the intensity of a stationary pairwise interaction point process in Section~\ref{sec:the intensity of a stationary pairwise interaction point process}. In Section~\ref{sec:numerical results}, the performance of the proposed scheme is compared numerically with the birth-death Metropolis-Hastings simulation and the perfect sampling. In the latter simulation,  point processes are generated by the dominated CFTP  algorithm.  Finally, we conclude the paper in Section~\ref{sec:conclusion}.

	\section{Monte Carlo methods for locally stable point processes}
	\label{sec:MC methods}
	Suppose that  $S$ is a bounded region on  $\mathbb R^d$.  
	A point process on $S$ is a set  of random number of points randomly located on $S$.
	A simple point process on $S$ is a homogeneous Poisson point process, which is a completely random point process on $S$ in the sense that the points of the process occur independently of one another.
	If $X$ is  the homogeneous Poisson point process with intensity $\rho \, (\rho > 0)$ on $S$, then for a subregion $A$ of $S$,  the number of points of $X$ belonging to $A$ follows the Poisson distribution with mean $\rho |A|$, where $|A|$ is the volume or the Lebesgue measure of $A$. For a collection of disjoint subregions of $S$, the number of points of $X$ belonging to a subregion is independent with those of other subregions.  In the case of $\rho = 1$, it is called the standard homogeneous Poisson point process.

	We call a set of finite points on $S$ a point pattern, and denote by $\mathcal{N}$ the collection of point patterns on $S$, i.e. $\mathcal{N} = \{x \subset S ; n(x) < \infty \}$, where $n(x)$ is the number of points of $x$. Then, $\mathcal{N}$ is also represented as follows:
	\begin{equation*}
		\mathcal{N} = \bigcup_{n=0}^{\infty} S^n, 
	\end{equation*}
	where $S^n = S \times \cdots \times S$, the $n$-fold Cartesian product of $S$ with itself, and $S^0$ is the empty set $\emptyset$.

	Let $X$ be  the homogeneous Poisson point process with intensity $\rho$ on $S$. 
	Then, the probability density function (pdf) of $X$ is as follows: 
	\begin{equation}
		\label{eq:pdf of homogeneous PPP}
		f_{P}(x;\rho) = \frac{ \rho^{n(x)} e^{- \rho |S|} }{n(x) !}, \quad x  \in \mathcal N.
	\end{equation}
	We can see that $f_P(x;\rho)$ does not depend on $x$, but depends on $n(x)$, and that 
	if the number of points of $X$ is given, then the points of $X$ are uniformly distributed on $S$, and their locations are mutually independent.

	The pdf $f_{P}(\cdot;\rho)$ in Eq~\eqref{eq:pdf of homogeneous PPP} induces a probability measure on $\mathcal N$, i.e. $\Pr (X \in dx) = f_P(x;\rho) dx$, $x \in \mathcal N$.  Then, it follows that for $\mathcal A \subset \mathcal{N}$, 
	\begin{equation}
		\label{eq:the measure induced by the standard homogeneous PP}
		\Pr(X \in \mathcal A) = \sum_{n=0}^\infty  \frac{\rho^{n} e^{- \rho|S|} }{n!} \int_{S^n}  I_{\mathcal A}(x) \, dx,    
	\end{equation}
	where  $I_{\mathcal A}(x)$ is the indicator function of $\mathcal A$. In the above equation, $\int_{S^0}  I_{\mathcal A}(x) \, dx$ is defined to be 1 if $\mathcal A$ contains the empty set.  We denote by Poi$(S,\rho)$ the probability measure defined in Eq~\eqref{eq:the measure induced by the standard homogeneous PP}, and call that a point process  $X$ follows Poi$(S,\rho)$ if $X$ satisfies Eq~\eqref{eq:the measure induced by the standard homogeneous PP}.

	For a point process $X$  on $S$,  the pdf of  $X$ with respect to Poi$(S,1)$ is defined as a function $f$ on $\mathcal N$ satisfying that  for $\mathcal A \subset \mathcal{N}$, 
	\begin{equation}
		\label{eq:the measure of a point process with respect to the standard homogeneous PP}
		\Pr(X \in \mathcal A) = \sum_{n=0}^\infty  \frac{ e^{- |S|} }{n!} \int_{S^n} I_{\mathcal A}(x) f(x) \, dx. 
	\end{equation}
	Due to Eq~\eqref{eq:the measure induced by the standard homogeneous PP}, 
	the above equation can be rewritten as follows: for $Y$ the standard homogeneous Poisson point process on $S$, 
	\begin{equation}
		\label{eq:the probability measure of X}
		\Pr(X \in \mathcal A) = E[I(Y \in \mathcal A)f(Y)], \quad  \mathcal A \subset \mathcal{N}.
	\end{equation}
	We assume that $f(x)$ has the following form: 
	\begin{equation} \label{def:nominal density}
		f(x) = \frac{1}{c_f} h(x), \; x \in \mathcal{N},
	\end{equation}
	where $h(\emptyset) = 1$ for the empty set  $\emptyset$. By substituting $\mathcal N$ for $\mathcal A$ in Eq~\eqref{eq:the probability measure of X}, we have that 
	\begin{equation} 
		\label{def:def of the normalizing constant}
		c_f = E[h(Y)].
	\end{equation}
	We also assume that the normalizing constant $c_f$ is unknown.

	The Papangelou conditional intensity of $X$  is defined as 
	\begin{equation} \label{def:papangelou conditional intensity}
		\lambda_f(x, \xi) = \frac{f(x \cup \xi)}{f(x)}, \quad x \in \mathcal{N}, \; \xi \in S \backslash x.
	\end{equation}
	$\lambda_f(x, \xi)$ can be considered as the ratio of the point patterns of  $x \cup \xi$ to $x$ in a huge collection of random copies of $X$ with pdf $f$.  $\lambda_f(x, \xi)$ represents the intensity that  $\xi$ will occur additionally given that the rest of $X$ is $x$.  Note that the Papangelou conditional intensity does not depend on $c_f$.

	If $\lambda_f(x, \xi)$ is bounded by an integrable function of $\xi$, i.e. for  a function $\phi$ defined on $S$ such that $\int_{S} \phi(\xi)d\xi <\infty$, 
	\begin{equation} \label{def:locally stable}
		\lambda_f(x, \xi) \le \phi(\xi) , \quad x \in \mathcal{N}, \; \xi \in S \backslash x, 
	\end{equation}
	then $f$ is said to be $\phi$-locally stable, and the point process with pdf $f$ is said to be locally stable. In order for the pdf $f$ in Eq~\eqref{def:nominal density}  to be well-defined, $c_f$ should be finite. In this case, it follows from Eq~\eqref{eq:the measure of a point process with respect to the standard homogeneous PP}  that $f$ induces a probability measure on $\mathcal N$ with respect to Poi$(S,1)$. A sufficient condition for $c_f$ to be finite is that $X$ is locally stable \citep[Chapter 6.1]{moller2003statistical}.

	A particular example of a locally stable point process is the stationary Strauss point process (Strauss \citeyearpar{strauss1975model}). Let $X$ be a finite stationary Strauss point process on $S$, and let $R$ be the range of interaction of $X$. Then, the pdf $X$ with respect to Poi$(S,1)$ is as follows: for $\beta > 0$, $0 \le \gamma \le 1$, and $R > 0$,
	\begin{equation} \label{eq:strauss point process}
		f(x; \beta, \gamma, R) \propto \beta^{n(x)} \gamma^{D(x)}, \quad x \in \mathcal{N},
	\end{equation}
	where $D(x) = \sum_{\{\xi, \eta\}\subseteq x}  I(\parallel \xi - \eta \parallel \le R)$. In the above pdf, $\beta$  controls how dense $X$ is, and $\gamma$ controls how regular $X$ is. If $\gamma = 0$, the process is called the hard-core process. If $\gamma = 1$, the process is equivalent to the homogeneous Poisson point process with intensity $\beta$.

	Applying the pdf~\eqref{eq:strauss point process} to Eq~\eqref{def:papangelou conditional intensity} gives that the Papangelou conditional intensity of $X$ is 
	\begin{equation} \label{eq:papangelou conditional intensity of strauss point process}
		\lambda_f(x, \xi) = \beta \gamma^{\sum_{\eta \in x} I(\parallel \xi - \eta \parallel \le R)}, \quad x \in \mathcal{N}, \xi \in S \backslash x.
	\end{equation}
	Since $0 \le \gamma \le 1$, it satisfies that $\lambda_f(x, \xi) \le \beta$.  It implies that a finite stationary Strauss point process is locally stable.
	
	\begin{figure}[t]
		\centering
		\begin{subfigure}{0.4\textwidth}
			\label{figure:realizations of SPP a}
			\centering
			\includegraphics[width=\textwidth]{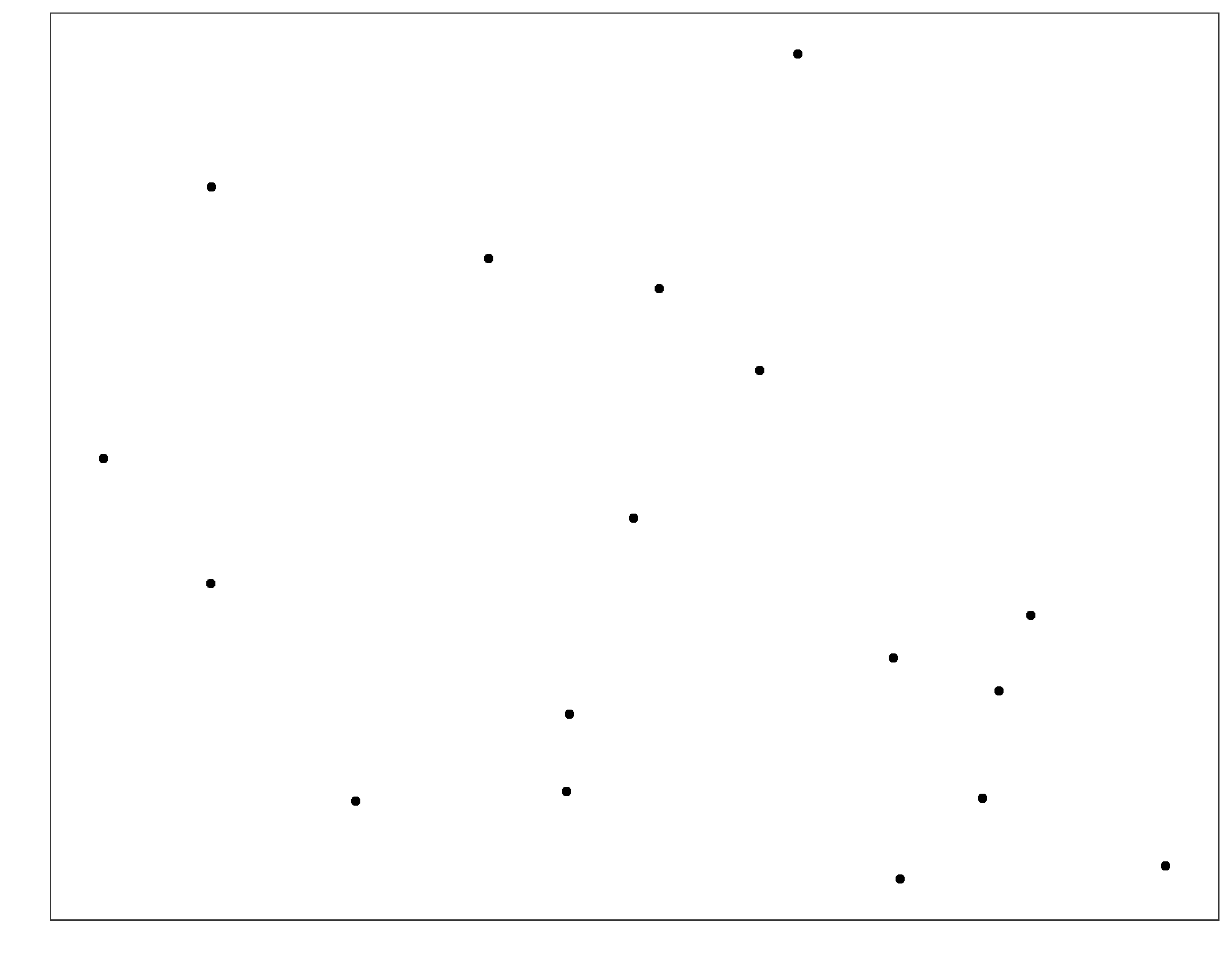}
			\caption{$\beta = 25$, $\gamma = 0.2$, $R=0.1$ }
		\end{subfigure}  \,
		\begin{subfigure}{0.4\textwidth}
			\centering
			\includegraphics[width=\textwidth]{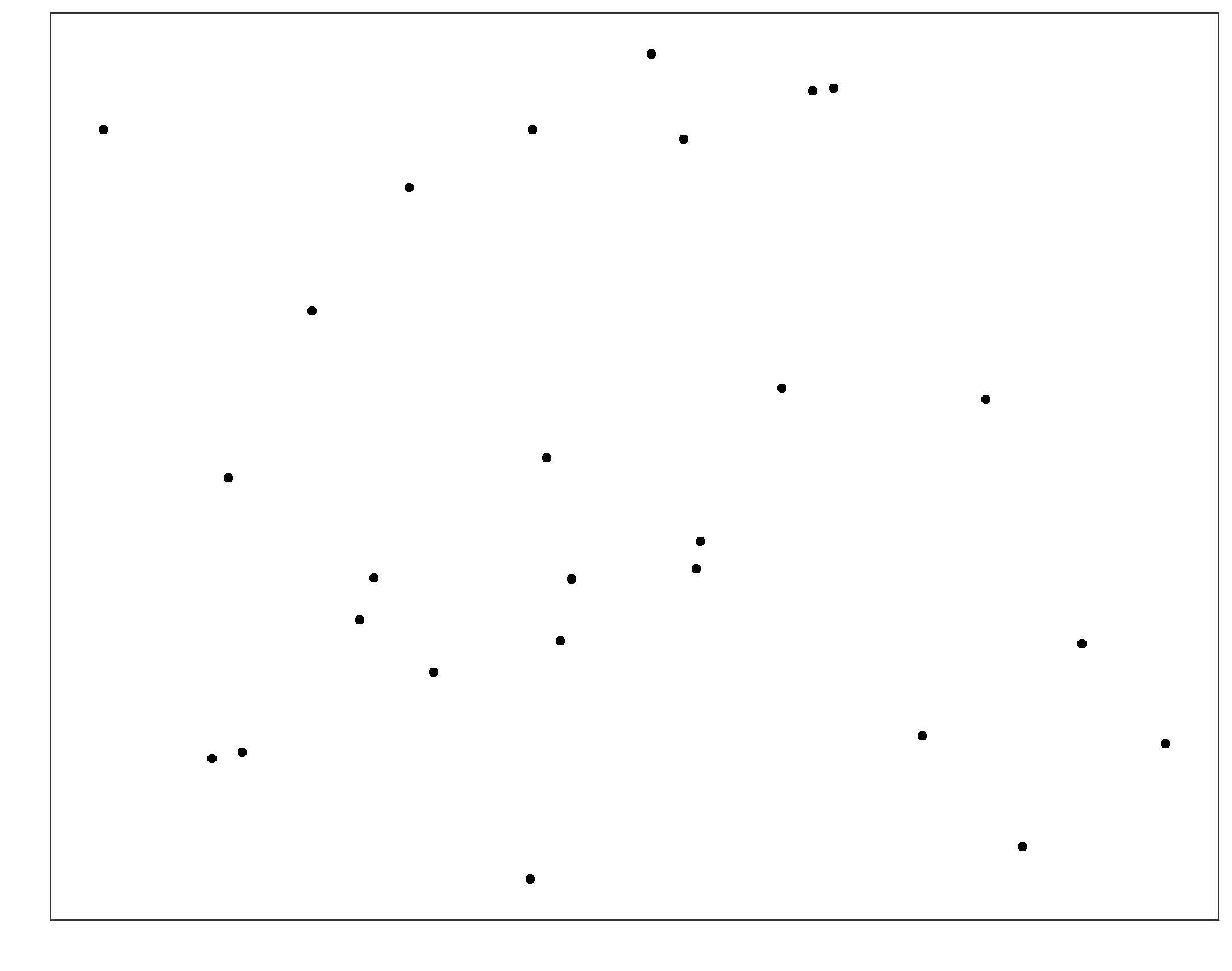}
			\caption{$\beta = 25$, $\gamma = 0.8$, $R=0.1$ }
		\end{subfigure} 
		\begin{subfigure}{0.4\textwidth}
			\centering 
			\includegraphics[width=\textwidth]{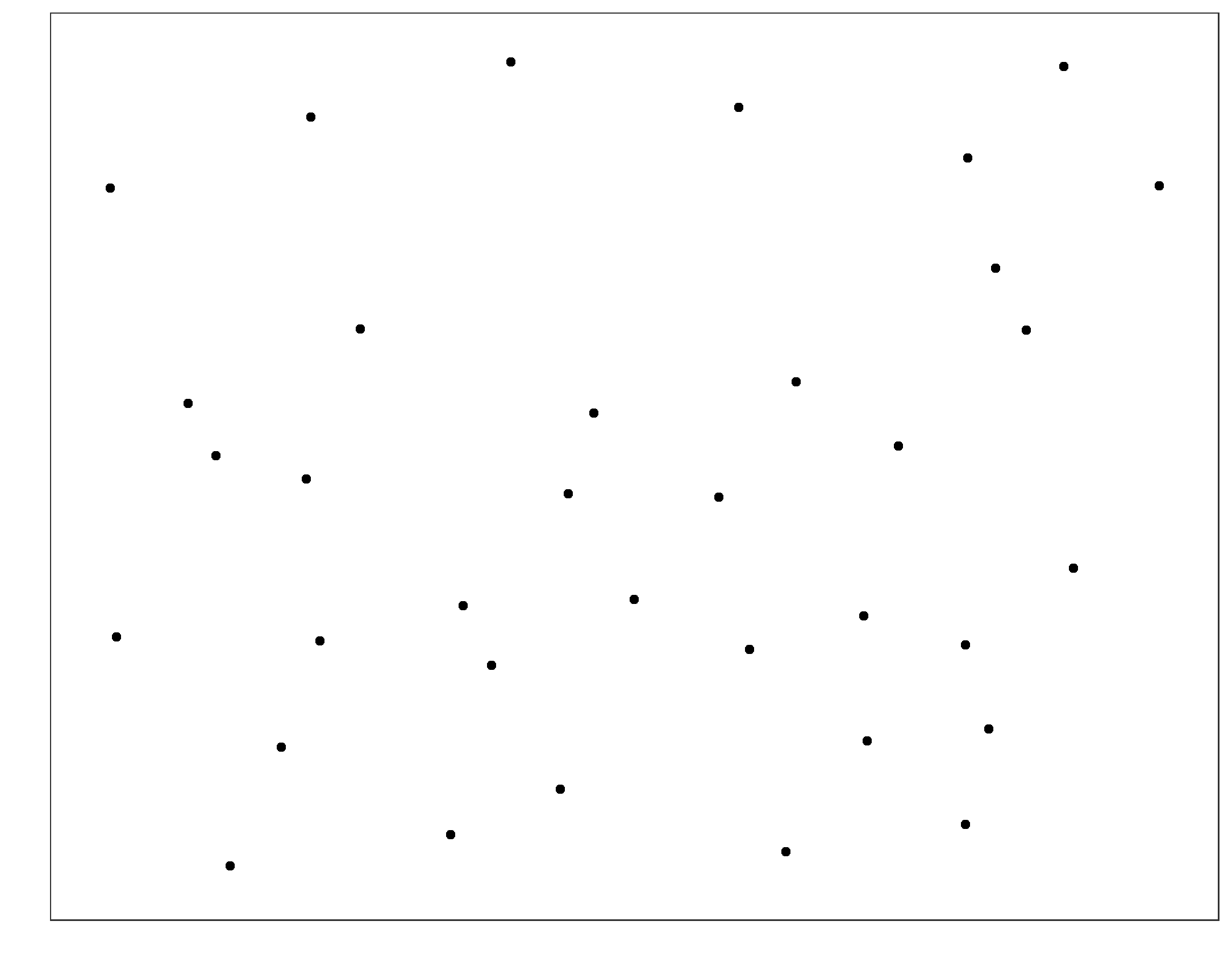}
			\caption{$\beta = 100$, $\gamma = 0.2$, $R=0.1$ }
		\end{subfigure}  \,
		\begin{subfigure}{0.4\textwidth}
			\centering
			\includegraphics[width=\textwidth]{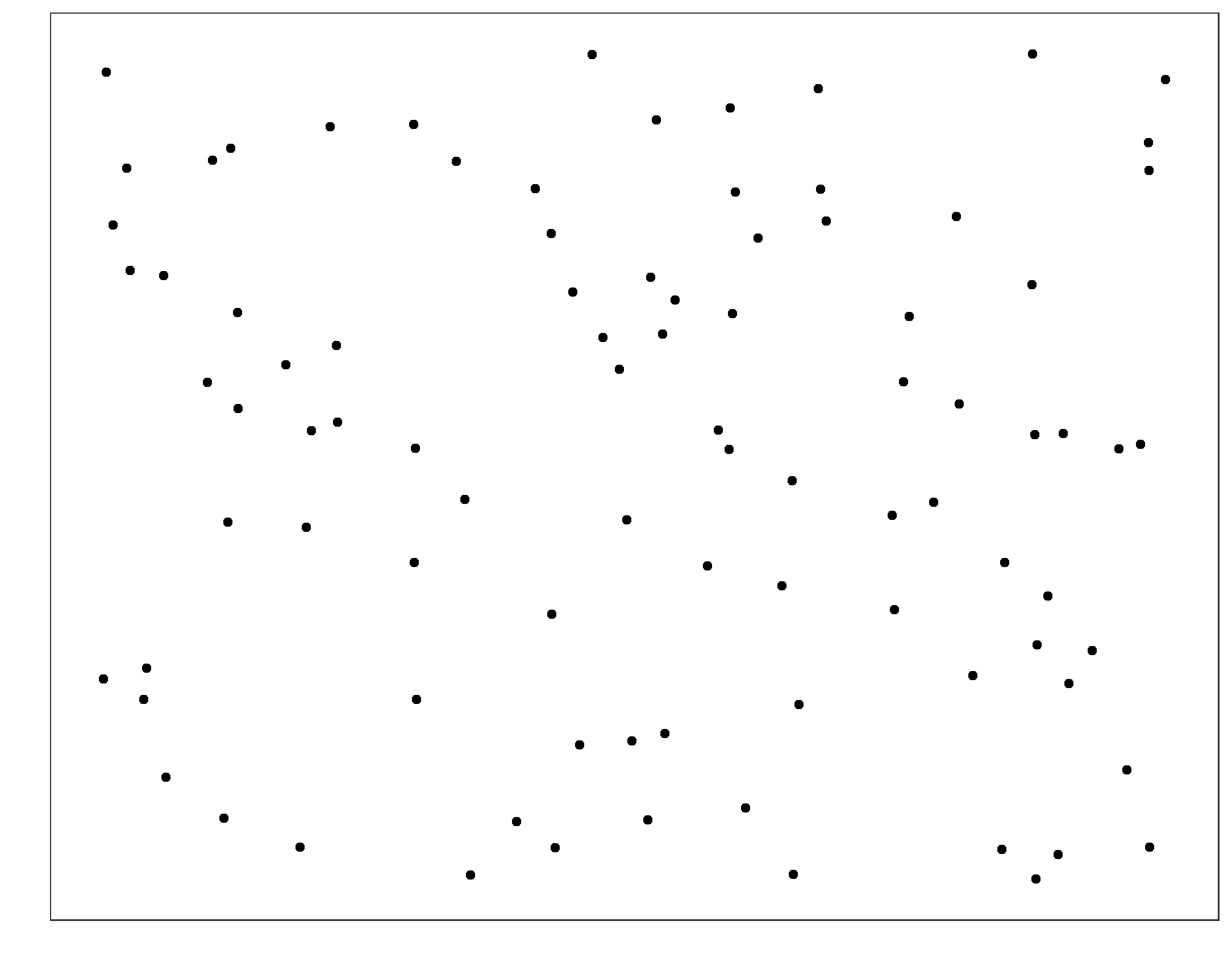}
			\caption{$\beta =100$, $\gamma = 0.2$, $R=0.025$ }
		\end{subfigure}
		\caption{Realizations of stationary Strauss point processes on the unit square with varying values of the parameters $\beta$, $\gamma$, $R$.}
		\label{fig:realizations of Strauss point processes}
	\end{figure}

	Figure~\ref{fig:realizations of Strauss point processes} illustrates the realizations of  stationary Strauss point processes on unit square  with varying values of the parameters $\beta$, $\gamma$, and $R$. The left two panels in the figure show that the more points occurred with increased value of $\beta$.
	By comparing top two panels, we can see that the distance between two neighboring points has decreased by increasing the value of $\gamma$ from $0.2$ to $0.8$.  In the two panels in the bottom of the figure, the range of interactions is $0.1$ for the left, and  $0.025$ for the right.  We can see that decreasing  the range of interaction also resulted in decreasing the distance between two neighboring points.

	We consider a locally stable point process $X$ on $S$. We assume that $f$, the pdf of $X$,  is $\phi$-locally stable for an integrable function $\phi$ on $S$. For a real-valued function $K$ defined on $\mathcal{N}$, we will denote by  $E_f[K(X)]$ the expected value of $K(X)$.  Let $\mu = E_f[K(X)]$.  
	It follows from Eq~\eqref{eq:the measure of a point process with respect to the standard homogeneous PP} that 
	\begin{equation*} \label{def:expectation explicit form}
		\mu = \sum_{n = 0}^{\infty} \frac{e^{-|S|}}{n !} \int_{S^n}  K(x)f(x)\, dx,
	\end{equation*}
	equivalently, 
	\begin{equation*}
		\mu = E[K(X)f(X)],
	\end{equation*}
	where $E[\cdot]$ is the expectation with respect to Poi$(S,1)$. 
	It follows from Eqs~\eqref{def:nominal density} and \eqref{def:def of the normalizing constant} that
	\begin{equation}
		\label{eq:a form of mu with respect to Poi(S,1)}
		\mu = \frac{E[K(X)h(X)]}{E[h(X)]}.
	\end{equation}
	In what follows, we assume that
	\begin{equation} \label{ineq:K square assumption}
		E_f[K^2(X)] < \infty,
	\end{equation}
	and that
	\begin{equation} \label{ineq:n square assumption}
		E_f[n^2(X)] <\infty.
	\end{equation}

	\subsection{Metropolis-Hastings algorithm for the estimation of $E_f[K(X)]$}
	\label{MH estimator}

	The birth-death Metropolis-Hastings algorithm is a type of Metropolis-Hastings algorithm for generating samples of a locally stable point process  \citep{moller2003statistical}. In this algorithm, samples of point processes  following a given pdf $f$ which is  $\phi$-locally stable are sequentially generated.

	At the initial step, we may generate a point pattern following a homogeneous Poisson point process with an appropriate intensity. Suppose that we have generated $X_n$, the point process at the $n$-th step. Then, $X_{n + 1}$ is generated by adding a point to $X_n$,  deleting a point from $X_n$, or keeping $X_n$ the same. The first is called a birth of a point, and the second a death of a point. Either of a birth or a death is proposed at  each step. If $X_n = \emptyset$, then only a birth proposal is possible. Let $p$ be the probability of proposing a birth. Suppose that $X_n= x$. If  a birth of a point  is proposed in this step, then we generate a point $\xi$ with pdf $q_b(x, \xi) = \phi(\xi)/c^*$, where $c^* = \int_S  \phi(\xi) \, d\xi$, as the candidate to be added. If a death of a point from $x$ is proposed, then we choose each point belonging to $x$ uniformly as the candidate to be deleted, i.e. $q_d(x, \eta) = 1/n(x)$ is the probability mass function for choosing a point $\eta \in x$ in a death proposal.

	In the case of the birth proposal, we compute the acceptance ratio of the birth of $\xi$ given by
	\begin{equation*}
		r_b(x, \xi) = \frac{\lambda_f(x, \xi)q_d(x \cup \xi)(1 - p)}{q_b(x, \xi) p},
	\end{equation*}
	and let $X_{n+1} = X_{n} \cup \xi$ with probability $\min \{1, r_b(x, \xi) \}$, and  $X_{n+1} = X_{n}$ with probability $1 - \min \{1, r_b(x, \xi) \}$.  In the other case, we compute  the acceptance ratio of the death of $\eta$ given by
	\begin{equation*}
		r_d(x, \eta) = \frac{1}{r_b(x\backslash\eta, \eta)},
	\end{equation*}
	and let $X_{n+1} = X_{n} \backslash \eta$ with probability $\min \{1, r_d(x, \eta) \}$, and  $X_{n+1} = X_{n}$ with probability $1 - \min \{1, r_d(x, \eta) \}$.

	When a sequence of samples $\{X_1, X_2, \ldots, X_n\}$  is  generated by the birth-death Metropolis-Hastings algorithm described above, they are dependent on each other, and do not follow $f$ exactly, but their stationary pdf is $f$. We remove some  initial samples from $\{X_1, X_2, \ldots, X_n\}$  that may not follow the stationary distribution.
	We apply subsampling from the remaining samples at fixed intervals to reduce the dependence between successive samples.

	Let $Y_1, Y_2, \dots, Y_n$ be the resulting samples. Then, they are near independent with stationary pdf $f$. The sample mean of $K(Y_1), K(Y_2), \dots$, $K(Y_n)$ is an estimator of $\mu$. We denote it by $\hat{\mu}_{\text{MH}}$, i.e,
	\begin{equation} \label{eq:MH estimator}
		\hat{\mu}_{\text{MH}} = \frac{1}{n} \sum_{i= 1}^n K(Y_i).
	\end{equation}
	We denote by MH method the sampling scheme described above.

	\subsection{Perfect sampling for the estimation of $E_f[K(X)]$}
	\label{CFTP estimator}
	The dominated coupling from the past (CFTP) algorithm is a perfect sampling for generating random samples of a locally stable point process with pdf $f$ on a bounded region $S$  \citep{kendall1999perfect, kendall2000perfect}. This algorithm constructs the backward step to generate a birth-death process and the forward step to update the dominating processes in reverse order of the birth-death process. To reduce the computational time, we employ the doubling scheme introduced by \cite{propp1996exact}. We denote the upper process as $U$ and the lower process as $L$, which are the dominating processes. Let $c^* = \int_S \phi(\xi)d\xi$ where $\phi(\cdot)$ is a $\phi$-function in Eq~\eqref{def:locally stable}.

	The  CFTP algorithm with doubling scheme proceeds as follows.
	\begin{enumerate}
		\item At the initial stage,  we generate $X_0 \sim \mbox{Poi}(S, \phi(\xi))$. If $X_0 = \emptyset$, then  we return $\emptyset$, and stop the procedure. Otherwise, at each stage $j=1,2, \ldots , $ we first execute the backward step,  and  then the forward step, which are described below.  Set $T_0 = 0$, $T_j = 2^{j – 1}$ for $j = 1, 2, \dots.$

		\item In the backward step of the $j$-th stage, we  initialize $X_{-T_{j - 1}}$ as the starting point process, and sequentially generate $X_t$ from time $-T_{j - 1} - 1$ to $-T_j$.   For $t \in \{-T_{j - 1}, \dots, -T_j + 1\}$, the death of a point in $X_t$ is proposed with probability $1 – c^*/(c^* + n(X_t))$. 
		In this case,  we randomly choose $\xi_{t – 1} \in X_t$, and  set $X_{t – 1} = X_t \backslash \xi_{t – 1}$.  We also assign a uniform random variable $R_{t - 1}$ on $[0, 1]$ to the point $\xi_{t – 1}$. Otherwise, we generate a point $\eta_{t – 1}$ from pdf $\phi(\eta)/c^*$,  and set $X_{t-1} = X_t \cup \eta_{t – 1}$.

		\item In the forward step of the $j$-th stage, we update sequentially the upper and the lower processes from time  $-T_j$ to $-1$. Let $U_{-T_j}^t$ and $L_{-T_j}^t$  be the upper and lower processes of $X_t$, respectively. At time $-T_{j}$, we initialize $U_{-T_j}^{-T_j} = X_{-T_j}$ and $L_{-T_j}^{-T_j} = \emptyset$.  If the death of a point was proposed at time $t \in \{-T_j, \dots, -1\}$ in the backward step, i.e., $X_{t + 1} \backslash \xi_{t} = X_{t}$, then the upper and lower processes are updated as follows:
		\begin{equation*}
			U_{-T_j}^{t + 1} = \left\{ \begin{array}{cc} U_{-T_j}^{t} \cup \xi_{t} & \mbox{if } R_{t} \le r_U \\
				U_{-T_j}^{t} & \mbox{otherwise} \end{array}\right.,
		\end{equation*}
		and
		\begin{equation*}
			L_{-T_j}^{t + 1} = \left\{ \begin{array}{cc} L_{-T_j}^{t} \cup \xi_{t} & \mbox{if } R_{t} \le r_L \\
				L_{-T_j}^{t} & \mbox{otherwise} \end{array}\right.,
		\end{equation*}
		where $r_U = \lambda_f{(L_{-T_j}^{t}, \xi_{t})}/\phi(\xi_{t})$, $r_L = \lambda_f{(U_{-T_j}^{t}, \xi_{t})}/\phi(\xi_{t})$, and $R_{t}$ is the uniform random variable assigned to  $\xi_{t}$ in the backward step.
		Otherwise (i.e., $X_{t + 1} \cup \eta_{t} = X_{t}$), the processes are updated as $U_{-T_j}^{t + 1} = U_{-T_j}^{t} \backslash \eta_{t}$ and $L_{-T_j}^{t + 1} = L_{-T_j}^{t} \backslash \eta_{t}$.

		\item If $U_{-T_j}^0 = L_{-T_j}^0$, then we return $U_{-T_j}^0$. Otherwise, we go to the next stage with setting $j \leftarrow j+1$. We repeat the above procedure until $U_{-T_j}^0 = L_{-T_j}^0$ is satisfied.
	\end{enumerate}

	Let $X_1, X_2, \dots, X_n$ be the  samples generated by applying the dominated CFTP algorithm with the doubling scheme. Then, an estimator of $\mu$ is given by
	\begin{equation} \label{eq:CFTP estimator}
		\hat{\mu}_{\text{CFTP}} = \frac{1}{n} \sum_{i = 1}^n K(X_i).
	\end{equation}
	The CFTP estimator in Eq~\eqref{eq:CFTP estimator} gives a  more  reliable estimate to $\mu$ than $\hat \mu_{\text{MH}}$ in terms of variance reduction. We denote by CFTP method the sampling scheme described above.

	\section{The proposed estimator}
	\label{sec:the proposed estimator}

	\subsection{Importance sampling for the estimation of $E_f[K(X)]$}
	Suppose that $g(x)$ is an importance sampling pdf of $X$, and that it dominates the nominal pdf $f$, i.e. $f(x) > 0$ implies $g(x) > 0$. It follows from Eq~\eqref{eq:a form of mu with respect to Poi(S,1)} that
	\begin{equation} 
		\mu = \frac{E_{g}[  K(X) w(X) ] }{E_{g}[w(X)]},
	\end{equation}
	where $w(x)$ is the unnormalized likelihood ratio of sample $x$, i.e.
	\begin{equation*}
		w(x) = \frac{h(x)}{g(x)}.
	\end{equation*}
	In Hesterberg \citeyearpar{hesterberg1995weighted}, if $X_1, \dots, X_n$ are samples from the importance sampling pdf~$g(x)$, then the self-normalized importance sampling estimator of $\mu$ is given as
	\begin{equation}
		\label{eq:the generic form of IS estimator}
		\hat{\mu}_{\text{IS}} = \frac{ \sum_{i = 1}^n K(X_i)w(X_i)}{\sum_{i = 1}^n w(X_i)} .
	\end{equation}
	Clearly, $\hat \mu_{\text{IS}}$ converges to $\mu$ with probability 1 as $n$ goes to the infinity (Owen~\citeyearpar{owen2013}).
	Under additional assumptions, the central limit theorem for $\hat{\mu}_{\text{IS}}$ was obtained by Geweke~\citeyearpar{geweke1989bayesian}.

	According to Oh and Berger \citeyearpar{oh1992adaptive} and Owen \citeyearpar{owen2013}, an approximate variance of $\hat{\mu}_{\text{IS}}$ is given by $\sigma^2/n$, where
	\begin{equation} \label{eq:variance for snise}
		\sigma^2 = \frac{E_g \left[ (K(X) - \mu)^2 w(X)^2 \right]}{E_g \left[w(X) \right]^2},
	\end{equation}
	or equivalently
	\begin{equation*}
		\sigma^2 = \frac{\mbox{Var}_g[K(X)w(X)] - 2\mu \mbox{Cov}_g[K(X)w(X), w(X)] + \mu^2\mbox{Var}_g[w(X)]}{E_g \left[w(X) \right]^2},
	\end{equation*}
	where $\mbox{Var}_g[\cdot]$ and $\mbox{Cov}_g[\cdot, \cdot]$ mean the variance and the covariance with respect to pdf $g(x)$, respectively.
	The value of $\sigma^2$ is estimated by
	\begin{equation} \label{eq:sample variance for snise}
		\begin{split}
			\hat{\sigma}^2 &= \frac{n \sum_{i = 1}^n (K(X_i) - \hat{\mu})^2 w(X_i)^2}{\left(\sum_{i = 1}^n w(X_i)\right)^2} \\
			&= \frac{\widehat{\mbox{Var}}_g[K(X)w(X)] - 2\hat{\mu} \widehat{\mbox{Cov}}_g[K(X)w(X), w(X)] + \hat{\mu}^2\widehat{\mbox{Var}}_g[w(X)]}{\bar{w}^2} ,
		\end{split}
	\end{equation}
	where
	\begin{equation*}
		\begin{split}
			\bar{w} &= \frac{1}{n} \sum_{i = 1}^n w(X_i),\\
			\widehat{\mbox{Var}}_g[K(X)w(X)] &= \frac{1}{n} \sum_{i = 1}^n K^2(X_i)w^2(X_i) - \hat{\mu}^2 \bar{w}^2, \\
			\widehat{\mbox{Cov}}_g[K(X)w(X), w(X)] &= \frac{1}{n} \sum_{i = 1}^n K(X_i)w^2(X_i) - \hat{\mu} \bar{w}^2, \\
			\widehat{\mbox{Var}}_g[w(X)] &= \frac{1}{n} \sum_{i = 1}^n w^2(X_i) - \bar{w}^2.
		\end{split}
	\end{equation*}
	
	Let $f^*(x)$  be the optimal importance sampling pdf for $\hat \mu_{\text{IS}}$ in terms of variance minimization.  Hesterberg \citeyearpar{hesterberg1995weighted} showed that
	\begin{equation*}
		f^*(x) \propto |K(x) - \mu|h(x), \quad x \in \mathcal{N}.
	\end{equation*}
	Since the pdf $f^*(x)$ includes the unknown value $\mu$, it can not be used as the importance sampling pdf.

	\subsection{An adaptive importance sampling estimator}
	\label{subsec: AIS estimator}
	In our proposal, we restrict the point process of the importance samples to the family of homogeneous Poisson point processes on $S$.  Suppose that $X$ follows Poi$(S, \rho)$. Then, it follows from Eqs~\eqref{eq:the measure induced by the standard homogeneous PP} and~\eqref{eq:the measure of a point process with respect to the standard homogeneous PP} that  the pdf of $X$ with respect to Poi$(S,1)$ is 
	\begin{equation} \label{def:homogeneous poisson point process}
		g(x; \rho) = \exp( (1 - \rho) |S|)\rho^{n(x)}, \quad x \in \mathcal{N}.
	\end{equation}

	The value of intensity $\rho$ minimizing the cross-entropy of $g(x; \rho)$ relative to $f^*(x)$ is found by solving the following optimization problem  \citep{rubinstein2016simulation}:
	\begin{equation}
		\label{eq:CE minimization method to find the sampling intensity}
		\maximize_{\rho} E_f[ |K(X) - \mu|\log g(X; \rho) ].
	\end{equation}
	From Eq~\eqref{def:homogeneous poisson point process}, the above problem is rewritten as
	\begin{equation*}
		\maximize_{\rho} E_f[|K(X) - \mu|]|S|(1 - \rho) + E_f[n(X)|K(X) - \mu|]\log \rho.
	\end{equation*}
	By differentiating the objective function of the above problem with respect to $\rho$,  we obtain the solution to the above problem as follows:
	\begin{equation*}
		\rho^* = \frac{1}{|S|}\frac{E_{f}\left[n(X) |K(X) -\mu|\right]}{E_{f}\left[|K(X) - \mu|\right]}.
	\end{equation*}

	Since $\mu$ is unknown, it is not an easy task to get a consistent estimator of $\rho^*$. If we knew the normalization constant $c_f$, then the optimal importance sampling pdf would be as follows \citep[Chapter 5]{rubinstein2016simulation}:
	\begin{equation*}
		f^*(x)  = \frac{|K(x)| h(x)}{c_f}, \quad x \in \mathcal{N}.
	\end{equation*}
	Then, the value of intensity $\rho$ minimizing the cross-entropy of $g(x; \rho)$ relative to $f^*(x)$ in the above equation is found by solving the following optimization problem instead of Eq~\eqref{eq:CE minimization method to find the sampling intensity}:
	\begin{equation*}
		\maximize_{\rho} E_f[ |K(X)|\log g(X; \rho) ].
	\end{equation*}
	In the same manner as the above,  it can be shown that the optimal intensity in this case is given by
	\begin{equation} 
		\label{eq:optimal rho}
		\rho' = \frac{1}{|S|}\frac{E_{f}\left[n(X) |K(X)|\right]}{E_{f}\left[|K(X) |\right]}.
	\end{equation}
	We consider $\rho'$ as a pseudo-optimal parameter  of the importance sampling pdf $g(x; \rho)$. Eqs (\ref{ineq:K square assumption}) and (\ref{ineq:n square assumption}) show the existence of $\rho'$.

	We propose an adaptive importance sampling to estimate $\mu$. In the proposed method, the value of $\mu$ is estimated iteratively, and the importance sampling pdf at each step is a homogeneous Poisson process whose intensity converges to a value close to  $\rho'$.  Let $\hat \rho_t$, $t=1,2, \ldots,$ be the estimator of $\rho'$ at the $t$-th step. Then, we generate a number of $n_t$ samples of point processes independently from the importance sampling pdf $g(x; \hat{\rho}_{t - 1})$ at the $t$-th step.  In order to get a  consistent estimator of $\mu$, we restrict the value of $\hat \rho_t$ to the interval $[m_{\rho}, M_{\rho}]$ for a sufficiently small $m_{\rho}$ and a sufficiently large  $M_{\rho}$. Initially, we choose $\hat{\rho}_0 \in [m_\rho, M_\rho]$.

	Let $X_1^{(t)}, \dots ,X_{n_t}^{(t)}$ be the generated samples at the $t$-th step. The unnormalized likelihood ratio of $X_i^{(t)}$ is given by
	\begin{equation*} 
		\label{eq:sample likelihood ratio}
		w_t(X_i^{(t)}) = \frac{h(X_i^{(t)})}{g(X_i^{(t)}; \hat{\rho}_{t - 1})}.
	\end{equation*}
	The estimator of $\mu$ at the $t$-th step is given by
	\begin{equation} 
		\label{eq:proposed estimator of mu}
		\hat{\mu}_{t} =  \frac{ \sum_{j = 1}^{t} \sum_{i = 1}^{n_{j}} K (X^{(j)}_i )w_{j} (X^{(j)}_i)}{\sum_{j = 1}^{t} \sum_{i = 1}^{n_{j}} w_{j}(X^{(j)}_i )}.
	\end{equation}

	If we set $\hat \rho_t$ to be the plug-in estimator of $\rho'$ given in  Eq~\eqref{eq:optimal rho}, then it cannot  be guaranteed that $\hat \rho_t \in [m_\rho, M_\rho]$.
	Instead, we define  $\tilde{n}(x)$ as the truncated value of $n(x)$ with lower bound of  $m_{\rho}|S|$ and upper bound of $M_{\rho}|S|$, and set
	\begin{equation} 
		\label{eq:proposed estimator of rho}
		\hat{\rho}_{t} = \frac{1}{|S|}  \frac{\sum_{j = 1}^{t}\sum_{i = 1}^{n_{j}} \tilde n(X_i^{(j)}) |K(X_i^{(j)}) | w_{j} (X^{(j)}_i )}{\sum_{j = 1}^{t} \sum_{i = 1}^{n_{j}} |K(X_i^{(j)}) |  w_{j}(X^{(j)}_i)}.
	\end{equation}
	Then, it can be easily checked that $\hat{\rho}_{t} \in [m_{\rho}, M_{\rho}]$.

	\begin{algorithm}[h]
		\caption{AIS for locally stable processes}
		\label{alg:adaptive importance sampling}
		\begin{algorithmic}[1]
			\Require{$h(x), K(x), m_{\rho}, M_{\rho}, \{n_1, n_2, \ldots \}$, $\eta_1, \eta_2$}
			\Ensure{$\hat{\mu}_t$}
			\State Choose $\hat{\rho}_0 \in [m_{\rho}, M_{\rho}]$
			\State Set $t = 1$
			\Repeat
			\State Generate $X_1^{(t)}, X_2^{(t)}, \dots, X_{n_t}^{(t)}$ independently from $g(x; \hat{\rho}_{t - 1})$
			\State Set $w_t(X_i^{(t)}) = \frac{h(X_i^{(t)})}{g(X_i^{(t)}; \hat{\rho}_{t - 1})}$ for $i = 1, 2, \dots, n_t$
			\State Compute
			$$\hat{\mu}_{t} =  \frac{\sum_{j = 1}^{t} \sum_{i = 1}^{n_{j}} K(X^{(j)}_i)w_j(X_i^{(j)})}{\sum_{j = 1}^{t} \sum_{i = 1}^{n_{j}} w_j(X_i^{(j)})}$$
			\State Set $\tilde{n}(X_i^{(t)}) = \median\{m_{\rho}|S|, n(X_i^{(t)}), M_{\rho}|S|\}$ for $i = 1, 2, \dots, n_t$
			\State Compute
			$$\hat{\rho}_{t} = \frac{1}{|S|} \frac{\sum_{j = 1}^{t} \sum_{i = 1}^{n_{j}} \tilde n(X_i^{(j)}) |K(X_i^{(j)}) | w_j(X_i^{(j)})}{\sum_{j = 1}^{t} \sum_{i = 1}^{n_{j}} |K(X_i^{(j)}) |w_j(X_i^{(j)})}$$
			\State Set $n^{(t)} = \sum_{j=1}^t n_j$
			\State Compute
			$$\hat{\sigma}_t^2 = \frac{n^{(t)}\sum_{j = 1}^t \sum_{i = 1}^{n_{j}} \left( K(X^{(j)}_i) - \hat{\mu}_t\right)^2w^2_j(X_i^{(j)})}{\left(\sum_{j = 1}^t \sum_{i = 1}^{n_{j}} w_j(X_i^{(j)})\right)^2}$$
			\State Set $t = t + 1$
			\Until{$\frac{\hat{\sigma}^{2}_t}{n^{(t)}\hat{\mu}^2_t} \le \eta_1$ and $\frac{|\hat{\rho}_t - \hat{\rho}_{t - 1}|}{\hat{\rho}_{t - 1}} \le \eta_2$}
			\State Return $\hat{\mu}_t$
		\end{algorithmic}
	\end{algorithm}

	We denote by $n^{(t)}$ the number of  samples of $X$ until the $t$-th step, i.e, $n^{(t)} = \sum_{j = 1}^t n_j$. It follows from Eq (\ref{eq:sample variance for snise}) that the approximate variance of $\hat{\mu}_t$ is estimated as $\hat{\sigma}^2_t/n^{(t)}$, where
	\begin{equation} \label{eq:variance for proposed estimator}
		\hat{\sigma}^2_{t} = \frac{n^{(t)} \sum_{j = 1}^{t} \sum_{i = 1}^{n_{j}}  \left( K( X^{(j)}_i) - \hat{\mu}_{t} \right)^2 w^2_{j}(X^{(j)}_i) }{\left(\sum_{j = 1}^{t} \sum_{i = 1}^{n_{j}} w_{j}(X^{(j)}_i)\right)^2}.
	\end{equation}
	The iterative estimation of $\mu$ and $\rho'$ continues until both the following conditions are satisfied;
	\begin{equation}
		\label{eq:stopping condition on hat mu}
		\frac{\hat{\sigma}_t^2}{n^{(t)}\hat{\mu}_t^2} \le \eta_1
	\end{equation}
	and
	\begin{equation}
		\label{eq:stopping condition on hat rho}
		\frac{|\hat{\rho}_t - \hat{\rho}_{t - 1}|}{\hat{\rho}_{t - 1}} \le \eta_2.
	\end{equation}

	The value of $\eta_1$ is determined as follows: for a sufficiently small $\epsilon > 0$,
	\begin{equation*}
		\eta_1 = \left(\frac{\epsilon}{z_{\alpha/2}} \right)^2,
	\end{equation*}
	where $z_{\alpha/2}$ is the $\left(1 - \frac{\alpha}{2}\right)$ quantile of the standard normal distribution.  The term in the left-hand side of Eq~\eqref{eq:stopping condition on hat mu} is the squared relative standard error of $\hat \mu_t$. We determine the values of $\epsilon$ and $\alpha$ appropriately so that if the squared relative standard error is sufficiently small, then the iteration stops. A small constant is assigned as the value of $\eta_2$ so that the sufficiently small change of $\hat \rho_t$ makes the iteration stop. The condition~\eqref{eq:stopping condition on hat mu} can be considered as the condition on the relative error of $\hat \mu_t$ as well as on the relative standard error. It will be explained in Section~\ref{subsec:stopping criteria} how the condition~\eqref{eq:stopping condition on hat mu} regulates the relative error of the final estimate to $\mu$, and how the values of $\epsilon$ and $\alpha$ are determined.  Why we should consider the condition~\eqref{eq:stopping condition on hat rho} as well as  the condition~\eqref{eq:stopping condition on hat mu} will be explained  also  in the section.
	Algorithm~\ref{alg:adaptive importance sampling} shows the procedure to obtain the proposed estimator, which will be denoted by AIS estimator.

	We assume that the sample sizes $n_1, n_2, \ldots,$  satisfy  the following three conditions:
	\begin{equation} 
		\label{ineq:the first condition of sample sizes}
		\sup_{t \ge 1} \frac{t \, n_t}{n^{(t)}} < \infty,
	\end{equation}
	\begin{equation}
		\label{ineq:the second condition of sample sizes}
		\sup_{t \ge 2}  \frac{1}{n^{(t)}} \sum_{j = 1}^{t - 1} j |n_{j + 1} - n_{j}| < \infty, 
	\end{equation}
	and 
	\begin{equation}
		\label{ineq:the third condition of sample sizes}
		\lim_{t \rightarrow \infty} \frac{1}{(n^{(t)})^2}\sum_{j = 1}^{t} n^2_j  = 0.  
	\end{equation}

	Under the conditions~\eqref{ineq:the first condition of sample sizes} and~\eqref{ineq:the second condition of sample sizes}, Etemadi~\citeyearpar[Theorem 1]{etemadi2006convergence} showed that for a sequence of random variables $\{Y_i, i=1,2, \ldots \}$, if $\sum_{i=1}^t Y_i / t$ converges to a random variable $Y_0$ almost surely, then 
	\begin{equation*}
		\lim_{t \rightarrow \infty} \frac{1}{n^{(t)}} \sum_{j=1}^t n_j Y_j  = Y_0, \quad a.s.
	\end{equation*}

	The above three conditions on the sample sizes combined with some other conditions on the functions $K$ and $\phi$ guarantee the almost sure convergence of $\{\hat \mu_t, t =1, 2, \ldots\}$ to the target value $\mu$, and also the asymptotic normality of it, which will be shown in the next section.

	According to Oh and Berger \citeyearpar{oh1992adaptive}, if the sample size in the first step is small, then the estimator $\hat{\mu}_t$ may have a large error, and it slows down the convergence of $\hat \mu_t$. Thus,  we also assume that $n_1$ is sufficiently large in Algorithm~\ref{alg:adaptive importance sampling}.

	\section{Asymptotic properties of the proposed estimator}
	\label{sec:asymptotic properties  of the proposed estimators}
	In this section, we prove that both $\{\hat{\mu}_t, t = 1, 2, \dots\}$ and $\{\hat{\rho}_t, t = 1, 2, \dots\}$ converge almost surely, also prove the asymptotic normality of $\{\hat{\mu}_t, t = 1, 2, \dots \}$. 
	In Section 4.1, we construct a martingale difference sequence of functionals of $\{X_{i}^{(j)}, j=1, 2, \ldots, i=1, \ldots, n_j\}$, and show the almost-sure convergence of it by applying the technique described in Feller~\citeyearpar[Theorem~3, p.243]{feller1introduction}.  Then, it gives that the denominator and the nominator of $\hat \mu_t$ converge to $E[K(X)h(X)]$ and $E[h(X)]$ almost surely,  respectively, which leads to the almost-sure convergence of $\{\hat{\mu}_t, t = 1, 2, \dots\}$ to $\mu$.  In the similar manner to this, we also show that $\hat{\rho}_t$ converges to a value close to $\rho'$ almost surely. 
	In Section 4.2, we construct a martingale triangular array of functionals of $\{X_{i}^{(j)}, j=1, 2, \ldots, i=1, \ldots, n_j\}$, and show that the array is a zero-mean square integrable martingale, and converges to a normal random in distribution. As an immediate corollary, the asymptotic normality of $\{\hat{\mu}_t, t = 1, 2, \dots,\}$ follows. Proofs for these results  are basically in the same line with \cite{oh1992adaptive}.

	\subsection{Convergence of the proposed estimator}\label{subsec:Convergence of the proposed estimator}
	Let  $\mathcal X_t = \{X^{(t)}_{i}, i= 1, \ldots, n_t \}$, $t=1,2, \ldots,$  be the $n_t$ number of  point processes generated at the $t$-th step in Algorithm~\ref{alg:adaptive importance sampling}.  We define $\mathcal{F}_t$ as the $\sigma$-algebra generated by $\{ \mathcal X_j, j=1,2, \ldots, t \}$. Then, $\{\mathcal{F}_t, t = 1, 2, \ldots\}$ is a filtration. We also define $\mathcal{F}_\infty$ as the smallest $\sigma$-algebra containing $\{\mathcal{F}_t, t = 1, 2, \ldots\}$. Let $\bm Y$ be the  stochastic process  $\{\mathcal X_1, \mathcal X_2, \ldots \}$ of  the generated  point processes. We define $\Omega$ as the set of the possible sample paths of $\bm Y$, and define $G$ as the probability measure induced by  $\bm Y$.
	Then, $(\Omega, \mathcal F_\infty, G)$ is a probability space.

	For a $\mathcal F_\infty$-measurable function $H$, we denote by $E_G[H(\bm Y)]$ the expectation of $H(\bm Y)$ with respect to measure $G$, and by $E_G[H(\bm Y)|\mathcal{F}_{t}]$ the conditional expectation  of $H(\bm Y)$ with respect to $\mathcal{F}_{t}$. Since $\hat \rho_{t}$ is a function of $\{ \mathcal X_1, \mathcal X_2, \ldots, \mathcal X_t \}$ and the stochastic behavior of $\mathcal X_{t+1}$ is completely determined by $\hat \rho_{t}$, $E_G[H(\mathcal X_{t+1})|\mathcal F_{t}]$ can be considered as the expectation with respect to pdf $g(x; \hat \rho_t)$, i.e. the expected value of $H(\mathcal X_{t+1})$ when $X^{(t+1)}_{i}$, $i= 1, \ldots, n_{t+1}$, follows $g(x; \hat \rho_t)$.  For sake of notation, we let  $E_G[H(\mathcal X_1)|\mathcal F_{0}]$ be the expectation with respect to $g(\cdot;\hat \rho_0)$.

	For the sample point processes  $\{X^{(t)}_{i}, i= 1, \ldots, n_t \}$, we define $\psi_t$ and $\Psi_t$  as follows: for $t = 1,2, \ldots ,$
	\begin{equation} \label{def:psi and Psi functional_1}
		\begin{split}
			\psi_t(X_i^{(t)}) &= K(X_i^{(t)})w_t(X_i^{(t)}) - E_G[K(X_i^{(t)})w_t(X_i^{(t)})|\mathcal{F}_{t - 1}], \\
			\Psi_t &= \sum_{i = 1}^{n_t} \psi_t(X^{(t)}_i).
		\end{split}
	\end{equation}
	Note that both $\psi_t$ and $\Psi_t$ are functionals depending on the function $K(x)$,
	and that $E_G[K(X_i^{(t)})w_t(X_i^{(t)})|\mathcal{F}_{t - 1}]$ is equal to  $E_{g_{t-1}}[K(X_i^{(t)})w_t(X_i^{(t)})]$, where $E_{g_{t-1}}[\cdot]$ denotes the expectation with respect to $g(x;\hat \rho_{t-1})$.
	Since $w_t(x) = h(x)/g(x; \hat{\rho}_{t - 1})$,  we have that
	\begin{equation*}
		E_G[K(X_i^{(t)})w_t(X_i^{(t)})|\mathcal{F}_{t - 1}] = E[K(X)h(X)].
	\end{equation*}
	Then, Eq (\ref{def:psi and Psi functional_1}) is rewritten as
	\begin{equation*} \label{eqs:psi and Psi functional_2}
		\begin{split}
			\psi_t(X_i^{(t)}) &= K(X_i^{(t)})w_t(X_i^{(t)}) - E[K(X)h(X)], \\
			\Psi_t&= \sum_{i = 1}^{n_t} K(X_i^{(t)})w_t(X_i^{(t)}) - n_tE[K(X)h(X)].
		\end{split}
	\end{equation*}

	Suppose that  for a non-negative valued function $\rho(\xi)$ on $S$, $\int_S \rho(\xi) d\xi < \infty$. Then, the pdf of the non-homogeneous Poisson point process with intensity function $\rho(\xi)$, $\xi \in S$, with respect to Poi$(S, 1)$ is as follows \citep{moller2003statistical}:
	\begin{equation} \label{def:non homogeneous poisson point process}
		p(x; \rho(\xi)) = \exp \left( |S| - \int_{S} \rho(\xi) d\xi \right) \prod_{\xi \in x} \rho(\xi), \quad x \in \mathcal{N}.
	\end{equation}
	For convenience of presentation, we denote by $q_i(x)$ the pdf $p\left (x,\frac{1}{m^i_{\rho}} \phi^{i + 1}(\xi) \right )$, and by $E_{q_i}[\cdot]$ the expectation with respect to $q_i(x)$ for $i = 1, 2, 3$.

	\begin{theorem} \label{thm:the almost sure convergence of  the averaged K(X)w(x)}
		Suppose that $\int_{S} \phi^2(\xi)d\xi < \infty$ and $E_{q_1}[K(X)^2] < \infty$. Then, for a point process $X$ following Poi$(S,1)$,
		\begin{equation}
			\label{eq:almost sure convergence of the averaged value of K(X)w(X)}
			\lim_{t \rightarrow \infty} \frac{1}{n^{(t)}} \sum_{j = 1}^t \sum_{i=1}^{n_j} K(X_i^{(j)})w_j(X_i^{(j)}) = E[K(X)h(X)],  \quad  a.s.
		\end{equation}
	\end{theorem}
	\begin{proof}
		We can see from the form of $\Psi_t$ given in Eq~\eqref{def:psi and Psi functional_1} that Eq~\eqref{eq:almost sure convergence of the averaged value of K(X)w(X)} is equivalent to
		\begin{equation*}
			\lim_{t \rightarrow \infty} \frac{1}{n^{(t)}} \sum_{j = 1}^t  \Psi_j = 0, \quad  a.s.
		\end{equation*}
		Eq (\ref{def:psi and Psi functional_1}) gives that
		\begin{equation} \label{eq:conditional expectation for psi}
			E_G[ \psi_t(X_i^{(t)}) | \mathcal{F}_{t - 1}] = 0, \quad i = 1, \ldots, n_t,
		\end{equation}
		and that
		\begin{equation*}
			E_G[ \Psi_t | \mathcal{F}_{t - 1}] = 0, \quad t=1,2, \ldots.
		\end{equation*}
		Then, due to Feller~\citeyearpar[Theorem~3, p.243]{feller1introduction}, it is sufficient to show that
		\begin{equation*}
			\sum_{t = 1}^{\infty}\frac{E_G[\Psi_t^2]}{(n^{(t)})^2}  < \infty.
		\end{equation*}

		We will first show that $E_G[\psi^2_t(X_i^{(t)})], \, t \geq 1, $ is bounded above by a constant.  It can be easily shown from the definition of $\psi_t(x)$ that
		\begin{equation}
			\label{eq:a bound on the conditional expectation on psi_t square}
			E_G [\psi_t^2(X_i^{(t)}) \big| \mathcal{F}_{t - 1} ] \leq E_G[K^2(X_i^{(t)})w^2_t(X_i^{(t)}) | \mathcal{F}_{t - 1}].
		\end{equation}
		It follows from Eqs (\ref{def:papangelou conditional intensity}) and (\ref{def:locally stable}) that
		\begin{equation} \label{ineq:upper for function h}
			h(x) \le \prod_{\xi \in x} \phi(\xi), \quad x \in \mathcal{N}.
		\end{equation}
		Note that $w_t(x) = h(x)/g(x; \hat{\rho}_{t - 1})$. Then, Eqs (\ref{def:homogeneous poisson point process}) and (\ref{ineq:upper for function h}) imply that
		\begin{equation*}
			w_t(x) < \exp\left(\hat{\rho}_{t - 1}|S| \right) \prod_{\xi \in x} \frac{\phi(\xi)}{\hat{\rho}_{t - 1}}.
		\end{equation*}
		Since $m_{\rho} \le \hat{\rho}_{t - 1} \le M_{\rho}$, we have that
		\begin{equation} \label{ineq:upper for weight}
			w_t(x) < \exp\left(M_{\rho}|S|\right) \prod_{\xi \in x} \frac{\phi(\xi)}{m_{\rho}}.
		\end{equation}
		Then, we have from Eq~\eqref{def:non homogeneous poisson point process}  that
		\begin{equation}
			\label{eq:a upper bound on w_t(x) h(x)}
			w_t(x) h(x) <  \exp\left(M_{\rho}|S| + \frac{1}{m_{\rho}} \int_S \phi^{2}(\xi)d\xi \right) q_1(x).
		\end{equation}
		Since $w^2_t(x) =  w_t(x) h(x)/g(x;\hat \rho_{t-1})$, we have that
		\begin{equation*}
			\begin{split}
				& E_G[K^2(X^{(t)}_i)w_t^2(X^{(t)}_i)|\mathcal{F}_{t - 1}] \\
				& < \exp\left(M_{\rho}|S| + \frac{1}{m_{\rho}} \int_S \phi^{2}(\xi)d\xi \right)  E_G\left [ \frac{K^2(X^{(t)}_i) q_1(X^{(t)}_i)}{g(X^{(t)}_i;\hat \rho_{t-1})} \big |\mathcal{F}_{t - 1} \right ].
			\end{split}
		\end{equation*}
		We obtain from the above equation that
		\begin{equation*}
			\label{ineq:upper for conditional expectation K square w square}
			E_G[K^2(X^{(t)}_i)w_t^2(X^{(t)}_i)|\mathcal{F}_{t - 1}]
			< \exp\left(M_{\rho}|S| + \frac{1}{m_{\rho}}\int_S \phi^{2}(\xi)d\xi\right) E_{q_1}\left[ K^2(X) \right].
		\end{equation*}
		We denote by $C$ the right hand side of the above equation.  Then, $C$ is finite from the assumptions on $\phi$ and $K$.
		Eq \eqref{eq:a bound on the conditional expectation on psi_t square} shows that
		for $i=1,2, \ldots, n_t$,
		\begin{equation}
			\label{ineq:upper for psi square}
			E_G[\psi_t^2(X^{(t)}_i)] < C.
		\end{equation}

		Since $\psi_t(X_1^{(t)}), \dots, \psi_t(X_{n_t}^{(t)})$ given $\hat{\rho}_{t - 1}$  are conditionally i.i.d. with mean zero, it follows  that
		\begin{equation*} \label{eq:conditional expectation for Psi square}
			E_G[\Psi_t^2 \big| \mathcal{F}_{t - 1} ] = n_t E_G[ \psi_t(X^{(t)})^2 \big| \mathcal{F}_{t - 1} ],
		\end{equation*}
		where $X^{(t)}$ is a point process following $g(x;\hat \rho_{t-1})$. Then, we obtain that $E_G[\Psi_t^2] = n_t E_G[ \psi_t(X^{(t)})^2]$, which implies that
		\begin{equation*}
			E_G[\Psi_t^2] <  n_t C.
		\end{equation*}
		Let $u_0 = \sup_{t \ge 1} t \, n_t /n^{(t)}$. Then, $n_t/n^{(t)} \le u_0/t$ for $t \geq 1$.  Since $n^{(t)} \geq t$, we have that 
		\begin{equation*}
			\frac{n_t}{(n^{(t)})^2} \le \frac{u_0}{t^2}, \quad t \geq 1. 
		\end{equation*}
		Condition~\eqref{ineq:the first condition of sample sizes}   implies that
		\begin{equation} 
			\label{ineq:upper for sample size}
			\sum_{t = 1}^{\infty} \frac{n_t}{(n^{(t)})^2} < \infty.
		\end{equation}
		Then, we can see that $\sum_{t = 1}^{\infty}E_G[\Psi_t^2] / (n^{(t)})^2$ is finite.
	\end{proof}

	By exploiting Theorem~\ref{thm:the almost sure convergence of  the averaged K(X)w(x)},  we can show the almost sure convergence of $\{\hat \mu_t, t = 1,2, \ldots \}$ in Eq~\eqref{eq:proposed estimator of mu} under the same condition as Theorem~\ref{thm:the almost sure convergence of  the averaged K(X)w(x)}.
	
	\begin{theorem} \label{thm:the almost sure convergence of mu_t}
		Suppose that $\int_{S} \phi^2(\xi)d\xi < \infty$ and $E_{q_1}[K(X)^2] < \infty$. Then,
		$$ \lim_{t \rightarrow \infty} \hat{\mu}_t = \mu,  \quad \mbox{a.s}.$$
	\end{theorem}
	\begin{proof}
		Letting $K(X) = 1$ in Theorem~\ref{thm:the almost sure convergence of  the averaged K(X)w(x)}, we obtain that
		\begin{equation} \label{eq:convergence for w}
			\lim_{t \rightarrow \infty} \frac{1}{n^{(t)}} \sum_{j = 1}^{t} \sum_{i = 1}^{n_j} w_j(X_i^{(j)}) = E[h(X)], \quad \mbox{a.s.}
		\end{equation}
		Theorem~\ref{thm:the almost sure convergence of  the averaged K(X)w(x)} and the above equation give that
		\begin{equation*}
			\lim_{t \rightarrow \infty} \frac{\sum_{j = 1}^{t} \sum_{i = 1}^{n_j} K(X_i^{(j)})w_j(X_i^{(j)})}{\sum_{j = 1}^{t} \sum_{i = 1}^{n_j} w_j(X_i^{(j)})} = \frac{E[K(X)h(X)]}{E[h(X)]}, \quad \mbox{a.s.}
		\end{equation*}
		The right-hand side of the above equation is equal to $\mu$ in Eq~\eqref{eq:a form of mu with respect to Poi(S,1)}, which completes the proof. 
	\end{proof}

	In the proposed adaptive importance sampling shown in Algorithm~\ref{alg:adaptive importance sampling},  the importance samples of $X$ generated at the $t$-th step are the homogeneous Poisson point processes with intensity $\hat \rho_t$ in Eq~\eqref{eq:proposed estimator of rho}.  As mentioned in Section~\ref{subsec: AIS estimator}, the pseudo-optimal value of the intensity is $\rho'$ in Eq~\eqref{eq:optimal rho}.  However, $\{\hat \rho_t, t = 1, 2, \ldots \}$ does not converge to $\rho'$, but converges to  a value close to it.  We define $\tilde \rho$ as follows:
	\begin{equation*}
		\tilde \rho = \frac{1}{|S|}  \frac{E_f[\tilde n(X)|K(X)|]}{E_f[|K(X)|]}.
	\end{equation*}
	The following theorem shows that $\hat \rho_t$  converges to $\tilde \rho$ as $t$ goes to the infinity.

	\begin{theorem} \label{thm:the almost sure convergence of rho_t}
		Under the same condition as Theorem~\ref{thm:the almost sure convergence of mu_t}, we have that
		\begin{equation*}
			\lim_{t \rightarrow \infty} \hat \rho_t =  \tilde \rho, \quad \mbox{a.s.}
		\end{equation*}
	\end{theorem}
	\begin{proof}
		Suppose that $\int_{S} \phi^2(\xi)d\xi < \infty$ and $E_{q_1}[K^2(X)] < \infty$.
		Since $\tilde n(x)$, $x \in \mathcal{N}$, is bounded above by $M_\rho$, we also have that
		$E_{q_1}[\tilde n^2(X) K^2(X)] < \infty$. By substituting $\tilde n(x)|K(x)|$ for $K(x)$ in Theorem~\ref{thm:the almost sure convergence of  the averaged K(X)w(x)}, we obtain  that
		\begin{equation*}
			\lim_{t \rightarrow \infty} \frac{1}{n^{(t)}} \sum_{j = 1}^t \sum_{i=1}^{n_j} \tilde n(X_i^{(j)}) |K(X_i^{(j)})|w_j(X_i^{(j)}) = E[\tilde n(X)|K(X)|h(X)],  \quad  a.s.
		\end{equation*}
		By substituting $|K(x)|$ for $K(x)$ in Theorem~\ref{thm:the almost sure convergence of  the averaged K(X)w(x)}, we obtain that
		\begin{equation*}
			\lim_{t \rightarrow \infty} \frac{1}{n^{(t)}} \sum_{j = 1}^t \sum_{i=1}^{n_j} |K(X_i^{(j)})|w_j(X_i^{(j)}) = E[|K(X)|h(X)],  \quad  a.s.
		\end{equation*}
		The above two equations give that
		\begin{equation*}
			\lim_{t \rightarrow \infty} \frac{1}{|S|}  \frac{\sum_{j = 1}^t \sum_{i=1}^{n_j} \tilde n(X_i^{(j)}) |K(X_i^{(j)})|w_j(X_i^{(j)})}{\sum_{j = 1}^t \sum_{i=1}^{n_j}|K(X_i^{(j)})|w_j(X_i^{(j)})} = \frac{1}{|S|} \frac{E[\tilde n(X)|K(X)|h(X)]}{E[|K(X)|h(X)]},  \quad  a.s.
		\end{equation*}
		Since the right-hand side in the above equation is equal to $\tilde \rho$, the proof is complete. 
	\end{proof}

	An upper bound on the difference between $\rho'$ and $\tilde \rho$ is obtained as follows:
	\begin{equation*}
		|\rho' -  \tilde \rho| \le \frac{1}{|S|}  \frac{E_f[| n(X) - \tilde n(X) | |K(X)|]}{E_f[|K(X)|]}.
	\end{equation*}
	Since we have chosen a sufficiently small $m_\rho$ and a sufficiently large $M_\rho$,  $\tilde n(X)$ and $n(X)$ are the same with a probability close to 1,  and $\tilde \rho$ may be close to $\rho'$. Thus, in Algorithm~\ref{alg:adaptive importance sampling}, the importance samples of $X$ are generated near optimally as steps go by.

	The importance sampling pdf $g(x; \rho)$ in Eq~\eqref{def:homogeneous poisson point process}  is  continuous with respect to $\rho$. Due to Theorem~\ref{thm:the almost sure convergence of rho_t},  we can see that  for $x \in \mathcal{N}$,
	\begin{equation*}
		\lim_{t \rightarrow \infty} g(x; \hat \rho_t) = g(x;\tilde \rho), \quad a.s.
	\end{equation*}
	The above equation implies that $X^{(t)}$, the importance sample generated from $g(x; \hat \rho_{t-1})$,  converges in distribution  as $t \rightarrow \infty$, and that $g(x;\tilde \rho)$ is the limiting pdf.  For a point process $X$ following $g(x;\tilde \rho)$ and  a function $H(x)$, $x \in \mathcal{N}$, we denote by $E_{\tilde{g}}[H(X)]$ the expected value of $H(X)$.

	\subsection{Asymptotic normality of the proposed estimator}	\label{subsec:Asymptotic normality of the proposed estimator}
	
	We have shown that $\hat \mu_t$ converged to $\mu$ almost surely as $t \rightarrow \infty$ in Theorem~\ref{thm:the almost sure convergence of mu_t}. In this subsection, we will show the asymptotic normality of $\{\hat \mu_t, t =1, 2, \ldots\}$.  Let $K^{\prime}(x) = a_1 + a_2K(x), \, x \in \mathcal{N}$, for real-valued constants $a_1$ and  $a_2$.  We define $\psi_{j}^{K^{\prime}}(X)$ as $\psi_{j}(X)$ with $K(X)$ being substituted with  $K^{\prime}(X)$.
	For a positive integer $\tau$, we define
	\begin{equation*}
		\begin{split}
			r_{j}^{(\tau)} = \frac{1}{\sqrt{n^{(\tau)}}} \sum_{i = 1}^{n_j} \psi_{j}^{K^{\prime}}(X_i^{(j)}), \quad j =1,2, \ldots, \tau,
		\end{split}
	\end{equation*}
	and
	\begin{equation*}
		S_{\tau, t} = \sum_{j = 1}^t r_{j}^{(\tau)}, \quad t = 1,2, \ldots, \tau.
	\end{equation*}

	\begin{theorem} \label{thm:zero mean square integrable martingale}
		Under the same condition as Theorem~\ref{thm:the almost sure convergence of mu_t}, we have that for a positive integer  $\tau$, $\{S_{\tau, t} , t = 1, 2, \dots, \tau \}$ is a zero-mean square integrable martingale with respect to the filtration $\{\mathcal{F}_t, t = 1, 2, \dots\}$.
	\end{theorem}
	\begin{proof}
		By substituting $\psi_t^{K^{\prime}}(x)$ for $\psi_t(x)$ in Eq (\ref{eq:conditional expectation for psi}), we have that $E_G[\psi_t^{K^{\prime}}(X^{(t)}_i)| \mathcal{F}_{t - 1}] = 0$ for $i=1, 2, \ldots, n_t$, which means that $E_G[r_{t}^{(\tau)} | \mathcal{F}_{t - 1}] = 0$ for $t=1,2, \ldots, \tau$. Then, it follows that for $t = 2,3, \ldots, $
		\begin{equation*}
			\begin{split}
				E_G[S_{\tau, t}|\mathcal{F}_{t - 1}] &= E_G\left[S_{\tau, t - 1} + r_{t}^{(\tau)} \big| \mathcal{F}_{t - 1} \right] \\
				&= S_{\tau, t - 1} + E_G\left[r_{t}^{(\tau)} \big| \mathcal{F}_{t - 1} \right]\\
				&= S_{\tau, t - 1}.
			\end{split}
		\end{equation*}
		Since $E_G[S_{\tau, 1}] = 0$, we have that
		\begin{equation*}
			E_G[S_{\tau, t}] = 0, \mbox{ for }t = 1, 2, \dots, \tau.
		\end{equation*}
		Thus, $\{S_{\tau, t}, t = 1, 2, \dots, \tau\}$ is a zero-mean martingale.

		Since $\psi_t^{K^{\prime}}(X_1^{(t)}), \dots, \psi_t^{K^{\prime}}(X_{n_t}^{(t)})$ given $\hat{\rho}_{t - 1}$ are i.i.d with mean zero, it follows that
		\begin{equation} \label{eqs:r square conditional F_{t-1}}
			\begin{split}
				E_G\left[  \left(r_{t}^{(\tau)}\right)^2 \Big| \mathcal{F}_{t - 1} \right] &= \frac{1}{n^{(\tau)}} \sum_{i = 1}^{n_t} E_G\left[\psi_t^{K^{\prime}}(X_{i}^{(t)})^2 | \mathcal{F}_{t - 1}\right] \\
				&= \frac{n_t}{n^{(\tau)}}E_G\left[\psi_t^{K^{\prime}}(X^{(t)})^2 | \mathcal{F}_{t - 1}\right],
			\end{split}
		\end{equation}
		which gives that
		\begin{equation*}
			E_G\left[\left(r_{t}^{(\tau)}\right)^2 \right] = \frac{n_t}{n^{(\tau)}} E_G\left[\psi_t^{K^{\prime}}(X^{(t)})^2 \right], \quad t = 1, 2, \ldots, \tau.
		\end{equation*}
		Eq (\ref{ineq:upper for psi square}) implies that
		\begin{equation} \label{ineq:upper for sum r square}
			\begin{split}
				E_G\left[ \left(r_{t}^{(\tau)}\right)^2 \right] \le  \frac{n_t}{n^{(\tau)}} C,
			\end{split}
		\end{equation}
		where $C = \exp\left(M|S| + \frac{1}{m_{\rho}}\int_S \phi^2(\xi) d\xi \right)(a_1^2 + 2a_1a_2 E_{q_1}[K(X)] + a_2^2E_{q_1}[K(X)^2]).$
		The conditional expectation of $S_{\tau, t}^2$ is represented as follows:
		\begin{equation*}
			\begin{split}
				E_G[S^2_{\tau, t}|\mathcal{F}_{t - 1}] &= E_G\left[\left(S_{\tau, t - 1} + r_{t}^{(\tau)} \right)^2 \Big| \mathcal{F}_{t - 1}\right] \\
				&= E_G[S_{\tau, t - 1}^2| \mathcal{F}_{t - 1}] + 2E_G\left[ S_{\tau, t - 1}r_{t}^{(\tau)}\Big| \mathcal{F}_{t - 1}\right] +  E_G\left[ \left( r_{t}^{(\tau)}\right)^2 \Big| \mathcal{F}_{t - 1}\right].
			\end{split}
		\end{equation*}
		Since $E_G\left[ S_{\tau, t - 1} r_{t}^{(\tau)} \big|\mathcal{F}_{t - 1}\right] = S_{\tau, t - 1}E_G\left[ r_{t}^{(\tau)} \big| \mathcal{F}_{t - 1}\right]$ and $E_G\left[ r_{t}^{(\tau)} \big|\mathcal{F}_{t - 1} \right] = 0$, the above equation is rewritten as
		\begin{equation}
			\label{eq:difference of conditional expectation of s square}
			E_G[S^2_{\tau, t}|\mathcal{F}_{t - 1}] = E_G[S_{\tau, t - 1}^2| \mathcal{F}_{t - 1}] + E_G\left[ \left( r_{t}^{(\tau)}\right)^2 \Big | \mathcal{F}_{t - 1}\right],
		\end{equation}
		which gives that
		\begin{equation*}
			E_G[S_{\tau, t}^2] = E_G[S_{\tau, t - 1}^2] + E_G\left[ \left(r_{t}^{(\tau)}\right)^2\right], \; t = 2, 3, \dots, \tau.
		\end{equation*}
		Note that $S_{\tau, 1} =  r_{1}^{(\tau)}$. Then, it follows from the above recurrence relation that
		\begin{equation*}
			E_G\left[S^2_{\tau, t}\right] = \sum_{j = 1}^t E_G\left[\left( r_{j}^{(\tau)}\right)^2  \right], \; t = 1, 2, \dots, \tau.
		\end{equation*}
		By applying Eq (\ref{ineq:upper for sum r square}) to the above equation, we obtain that
		\begin{equation*}
			E_G\left[S^2_{\tau, t}\right] \le  C, \; t = 1, 2, \dots, \tau,
		\end{equation*}
		which completes the proof.
	\end{proof}

	In our proposed method, the importance sampling pdf of $X$ at each step converges to $g(x; \tilde \rho)$, the pdf of the homogeneous Poisson point process with intensity $\tilde \rho$. Let $\hat \mu_{\tilde \rho}$ be the importance sampling estimator of $\mu$ when the importance sampling pdf  of $X$ is equal to $g(x; \tilde \rho)$. In this case, the unnormalized likelihood ratio is given by $\tilde w(x) = h(x)/g(x;\tilde \rho)$. The exact form of $\hat \mu_{\tilde \rho}$ is given by  substituting $\tilde w(x)$  for $w(x)$ in Eq~\eqref{eq:the generic form of IS estimator}.
	An approximate variance of $\hat \mu_{\tilde \rho}$ can be found from Eq~\eqref{eq:variance for snise}.  Since the unnormalized likelihood ratio is $\tilde w(x)$,  the approximate variance of $\hat \mu_{\tilde \rho}$ is as follows:
	\begin{equation}
		\label{eq:the limit of sigma square}
		\tilde \sigma^2 =  \frac{E_{\tilde{g}}[(K(X) - \mu)^2\tilde w^2(X)]}{E_{\tilde{g}}[\tilde w(X)]^2}.
	\end{equation}
	We let
	\begin{equation} \label{eqs:optimal variance and covariance}
		\begin{split}
			\tilde \sigma_1^2 &= E_{\tilde{g}}[\tilde w(X)^2] - E_{\tilde{g}}[\tilde w(X)]^2, \\
			\tilde \sigma_{12} &= E_{\tilde{g}}[K(X)\tilde w^2(X)] - E_{\tilde{g}}[K(X) \tilde w(X)]E_{\tilde{g}}[\tilde w(X)],\\
			\tilde \sigma_{2}^2 &= E_{\tilde{g}}[K^2(X)\tilde w^2(X)] - E_{\tilde{g}}[K(X)\tilde w(X)]^2.
		\end{split}
	\end{equation}
	Then, it can be easily checked that
	\begin{equation}
		\label{eq:another form of the limit of sigma square}
		\tilde \sigma^2 = \frac{\mu^2 \tilde \sigma_1^2 -2\mu\tilde \sigma_{12} + \tilde \sigma_{2}^2}{E_{\tilde{g}}[\tilde w(X)]^2}.
	\end{equation}

	\begin{theorem} \label{thm:asymptotic normality for S}
		Suppose that $\int_S \phi^4(\xi) d\xi < \infty$, $E_{q_i}[K(X)^{i + 1}] < \infty$ for $i = 1, 2, 3$. Then, we have that for arbitrary constants $a_1, a_2$,
		$$S_{t,t} \rightarrow N(0, a_1^2\tilde \sigma_1^2 + 2a_1 a_2\tilde \sigma_{12} + a_2^2 \tilde \sigma_2^2) \quad in \; dist,$$
		where $N(\mu, \sigma^2)$ is the normal distribution with mean $\mu$ and variance $\sigma^2$.
	\end{theorem}

	\begin{proof}
		It follows from the definition of $\psi_t^{K^{\prime}}(X^{(t)}_i)$ that for $i=1,2, \ldots, n_t$,
		\begin{equation}
			\label{eq:conditional expectation of psi_t^K'}
			\begin{split}
				&E_G[\psi_t^{K^{\prime}}(X^{(t)}_i)^2 | \mathcal{F}_{t - 1}] \\
				&= E_G[K^{\prime}(X^{(t)}_i)^2w^2_t(X^{(t)}_i)|\mathcal{F}_{t - 1}] - E_G[K^{\prime}(X^{(t)}_i)w_t(X^{(t)}_i)|F_{t - 1}]^2.
			\end{split}
		\end{equation}
		We have that
		\begin{equation*}
			\begin{split}
				E_G[K^{\prime}(X^{(t)}_i)^2w^2_t(X^{(t)}_i)|\mathcal{F}_{t - 1}]
				&= E_G\left [K'(X^{(t)}_i)^2 \frac{h^2(X^{(t)}_i)}{g^2(X^{(t)}_i;\hat \rho_{t-1})} |\mathcal{F}_{t - 1} \right ] \\
				&=  E\left [K'(X)^2 \frac{h^2(X)}{g(X;\hat \rho_{t-1})} \right].
			\end{split}
		\end{equation*}

		Since $g(x; \rho)$ in Eq (\ref{def:homogeneous poisson point process}) is a continuous function of $\rho$, and $\hat \rho_t$ converges to $\tilde \rho$ almost surely, we have that  for $x \in \mathcal{N}$,
		\begin{equation*}
			\lim_{t \rightarrow \infty} K'(x)^2 \frac{h^2(x)}{g(x;\hat \rho_{t-1})} = K'(x)^2 \frac{h^2(x)}{g(x;\tilde \rho)}, \quad a.s.
		\end{equation*}
		We define $\zeta(x)$ as follows:
		\begin{equation*}
			\zeta(x) = \exp \left ( M_\rho |S| + \frac{1}{m_\rho} \int_S \phi^2(\xi) d\xi \right )K'(x)^2 q_1(x), \quad x \in \mathcal{N}.
		\end{equation*}
		Eq~\eqref{eq:a upper bound on w_t(x) h(x)} shows that  for $t=1,2, \ldots,$
		\begin{equation*}
			K'(x)^2 \frac{h^2(x)}{g(x;\hat \rho_{t-1})} < \zeta(x), \quad x \in \mathcal{N}.
		\end{equation*}

		We have from the assumptions of the theorem that both $\int_S \phi^2(\xi) \, d\xi$
		and $E_{q_1}[K^2(X)]$ are finite, which gives that $E[\zeta(X)]$ is also finite.
		Then, it follows from the dominated convergence theorem that
		\begin{equation*}
			\lim_{t \rightarrow \infty} E\left [K'(X)^2 \frac{h^2(X)}{g(X;\hat \rho_{t-1})} \right ] = E\left [K'(X)^2 \frac{h^2(X)}{g(X;\tilde \rho)} \right ], \quad a.s.,
		\end{equation*}
		equivalently, for $i=1,2,\ldots, n_t$,
		\begin{equation}
			\label{eq:convergence of the conditional expectation of  K'(X)w_t(X)}
			\lim_{t \rightarrow \infty} E_G[K^{\prime}(X^{(t)}_i)^2w^2_t(X^{(t)}_i)|F_{t - 1}]  = E_{\tilde g} [K'(X)^2 \tilde w^2(X) ], \quad a.s.
		\end{equation}
		Note that $E_G[K^{\prime}(X^{(t)}_i)w_t(X^{(t)}_i)|F_{t - 1}] = E[K^{\prime}(X)h(X)]$, which gives that
		\begin{equation*}
			E_G[K^{\prime}(X^{(t)}_i)w_t(X^{(t)}_i)|F_{t - 1}] = E_{\tilde g}[K^{\prime}(X)\tilde w(X)].
		\end{equation*}
		Then, it follows from Eq~\eqref{eq:conditional expectation of psi_t^K'} and the above two equations that
		\begin{equation*}
			\lim_{t \rightarrow \infty} E_G[\psi_t^{K^{\prime}}(X^{(t)}_i)^2 | \mathcal{F}_{t - 1}]
			= E_{\tilde g}[K'(X)^2 \tilde w^2(X) ] -  E_{\tilde g}[K^{\prime}(X)\tilde w(X)]^2, \quad a.s.
		\end{equation*}

		Due to Eq~\eqref{eqs:optimal variance and covariance}, the right-hand side of the above equation is computed to be $a_1^2\tilde \sigma_1^2 +2a_1a_2\tilde \sigma_{12} + a_2^2  \tilde \sigma_2^2$. Then, we obtain that
		\begin{equation*}
			\lim_{t \rightarrow \infty} \frac{1}{t} \sum_{j = 1}^t E_G[\psi_j^{K^{\prime}}(X^{(j)})^2 | \mathcal{F}_{j - 1}] = a_1^2\tilde \sigma_1^2 +2a_1a_2\tilde \sigma_{12} + a_2^2  \tilde \sigma_2^2, \quad \mbox{a.s.}
		\end{equation*}
		Eqs~\eqref{eq:difference of conditional expectation of s square} and~\eqref{eqs:r square conditional F_{t-1}} give that
		\begin{equation*}
			\begin{split}
				E_G\left[S_{t, t}^2 \big| \mathcal{F}_{t - 1}\right] &= \sum_{j = 1}^t E_G\left[\left(r_j^{(t)}\right)^2 \big|\mathcal{F}_{t - 1}\right] \\
				&= \frac{1}{n^{(t)}} \sum_{j = 1}^t n_j E_G\left[\psi_j^{K^{\prime}}(X^{(j)})^2 \big| \mathcal{F}_{j - 1}\right].
			\end{split}
		\end{equation*}
		Due to Etemadi \citeyearpar[Theorem 1]{etemadi2006convergence}, the above two equations imply that
		\begin{equation*} \label{eq:convergence for sum of conditional expectation r square}
			\lim_{t \rightarrow \infty} E_G\left[ S_{t, t}^2 \big| \mathcal{F}_{t - 1} \right] = a_1^2\tilde \sigma_1^2 +2a_1a_2\tilde \sigma_{12} + a_2^2 \tilde \sigma_2^2, \quad \mbox{a.s.}
		\end{equation*}

		Note that $\{S_{\tau,t}, t = 1,2, \ldots, \tau\}$ is a square integrable martingale for given $\tau$.  Then, by Hall et al. \citeyearpar[Corollary 3.1, p.58]{hall2014martingale}, it is sufficient to show that
		\begin{equation*}
			\lim_{t \rightarrow \infty} \sum_{j = 1}^t E_G\left[\left( r_{j}^{(t)} \right)^4 \Big| \mathcal{F}_{j - 1}\right] = 0,
		\end{equation*}
		for the completion of the proof.
		Since $E_G[  \psi_j^{K^{\prime}}(X_i^{(j)})| \mathcal{F}_{j - 1}] = 0$,  we have that
		\begin{equation*}
			\begin{split}
				\sum_{j = 1}^t E_G\left[\left( r_{j}^{(t)} \right)^4 \Big| \mathcal{F}_{j - 1}\right] &= \frac{1}{(n^{(t)})^2}  \sum_{j = 1}^t E_G\left[\left( \sum_{i = 1}^{n_j} \psi_j^{K^{\prime}}(X_i^{(j)}) \right)^4 \Bigg| \mathcal{F}_{j - 1}\right] \\
				&= \frac{1}{(n^{(t)})^2} \sum_{j = 1}^t  \Bigg( n_j E_G\left[ \psi_j^{K^{\prime}}(X^{(j)})^4 \big| \mathcal{F}_{j - 1} \right] \\
				& \qquad + 6n_j(n_j - 1) E_G\left[ \psi_j^{K^{\prime}}(X^{(j)})^2 \big| \mathcal{F}_{j - 1} \right]^2\Bigg).
			\end{split}
		\end{equation*}

		By applying the Jensen's inequality to $E_G\left[ \psi_j^{K^{\prime}}(X^{(j)})^2 \big| \mathcal{F}_{j - 1}\right]^2$, we obtain that
		\begin{equation} \label{eq:upper for sum of conditional expectation r 4 power}
			\sum_{j = 1}^t E_G\left[\left(r_{j}^{(t)} \right)^4 \Big| \mathcal{F}_{j - 1}\right] < \frac{6}{  (n^{(t)})^2} \sum_{j = 1}^t n_j^2 E_G\left[ \psi_j^{K^{\prime}}(X^{(j)})^4 \big| \mathcal{F}_{j - 1} \right].
		\end{equation}
		Eqs (\ref{ineq:upper for function h}) and (\ref{ineq:upper for weight}) imply that
		\begin{equation*}
			\begin{split}
				E_G\left[\psi_j^{K^{\prime}}(X^{(j)})^4|F_{j - 1}\right] &= E_G\left[ \left( K^{\prime}(X^{(j)})w_j(X^{(j)}) - E[K^{\prime}(X)h(X)]\right)^4 \Big| \mathcal{F}_{j - 1} \right] \\
				& \le \exp\left(3M_{\rho}|S| + \frac{1}{m_{\rho}^3} \int_{S} \phi^4(\xi)d\xi\right) E_{q_3}[K^{\prime}(X)^4] \\
				& + 4 \left| \exp\left(2M_{\rho}|S| + \frac{1}{m_{\rho}^2} \int_{S} \phi^3(\xi)d\xi\right) E_{q_2}[K^{\prime}(X)^3] E[K^{\prime}(X)h(X)] \right| \\
				& + 6 \exp\left(M_{\rho}|S| + \frac{1}{m_{\rho}} \int_{S} \phi^2(\xi)d\xi\right) E_{q_1}[K^{\prime}(X)^2] E[K^{\prime}(X)h(X)]^2 \\
				& + 5E[K^{\prime}(X)h(X)]^4.
			\end{split}
		\end{equation*}
		We denote by $C$ the right hand side of the above inequality.
		Then, $C$ is finite from the assumption of the theorem.
		Eq (\ref{eq:upper for sum of conditional expectation r 4 power}) gives that
		\begin{equation*}
			\sum_{j = 1}^t E_G\left[\left(  r_{j}^{(t)} \right)^4 \Big| \mathcal{F}_{j - 1}\right] <   \frac{6 C }{(n^{(t)})^2} \sum_{j = 1}^t n_j^2.
		\end{equation*}
		Applying condition~\eqref{ineq:the third condition of sample sizes} to the above inequality completes the proof.
	\end{proof}

	Now, we obtain the asymptotic normality of $\{\hat{\mu}_t, t = 1, 2, \ldots \}$ from Theorem~\ref{thm:asymptotic normality for S} as follows.

	\begin{corollary} \label{cor:the asymptotic normality of mu_t}
		Under the same conditions as Theorem \ref{thm:asymptotic normality for S}, we have that
		\begin{equation*}
			\sqrt{n^{(t)}} (\hat{\mu}_t - \mu) \rightarrow N(0, \tilde \sigma^2), \quad \mbox{in dist}.
		\end{equation*}
	\end{corollary}
	\begin{proof}
		Let $K^{\prime}(x) = K(x) - \mu$. Then, $E_f[K^{\prime}(X)] = 0$, which gives that
		\begin{equation*}
			E_G[K^{\prime}(X^{(t)})w_t(X^{(t)})|\mathcal{F}_{t - 1}] = 0.
		\end{equation*}
		It follows from Eq (\ref{def:psi and Psi functional_1})  that
		\begin{equation*}
			\psi_t^{K^{\prime}}(X_i^{(t)}) = (K(X_i^{(t)}) - \mu )w_t(X_i^{(t)}).
		\end{equation*}
		Theorem \ref{thm:asymptotic normality for S} says that
		\begin{equation*} \label{eq:asymptotic normality for kw}
			\begin{split}
				\frac{1}{\sqrt{n^{(t)}}} \sum_{j = 1}^t\sum_{i = 1}^{n_j} (K(X_i^{(t)}) - \mu )w_t(X_i^{(t)}) \rightarrow N(0, \mu^2\tilde \sigma_1^2 -2\mu\tilde \sigma_{12} + \tilde \sigma_2^2), \quad in \; dist.
			\end{split}
		\end{equation*}
		Dividing the left hand side of the above equation  by $\sum_{j = 1}^t \sum_{i = 1}^{n_j} w_j(X_i^{(j)})/n^{(t)}$ gives $\sqrt{n^{(t)}} (\hat{\mu}_t - \mu)$. Eq~\eqref{eq:convergence for w} says that $\sum_{j = 1}^t \sum_{i = 1}^{n_j} w_j(X_i^{(j)})/n^{(t)}$ converges to $E_{\tilde g}[\tilde w(X)]$ almost surely. We have from the Slutsky theorem that
		\begin{equation*}
			\sqrt{n^{(t)}}\left(\hat{\mu}_t - \mu \right) \rightarrow N\left(0, \frac{\mu^2\tilde \sigma_1^2 -2\mu\tilde \sigma_{12} + \tilde \sigma_2^2}{E_{\tilde g}[\tilde w(X)]^2}\right), \quad in \; dist.
		\end{equation*}
		The variance of the above normal distribution is equal to $\tilde \sigma^2$ according to Eq~(\ref{eq:another form of the limit of sigma square}).
	\end{proof}

	Given the importance samples of $X$ generated until the $t$-th step in Algorithm~\ref{alg:adaptive importance sampling}, the approximate variance of $\hat \mu_t$  is computed from Eq~\eqref{eq:variance for proposed estimator}. The following theorem shows that  the approximate variance of $\hat \mu_t$ converges to $\tilde  \sigma^2$ almost surely.

	\begin{theorem} \label{thm:almost surely converge to sigma^2}
		Under the same conditions as Theorem \ref{thm:asymptotic normality for S}, we have that
		\begin{equation*}
			\lim_{t \rightarrow \infty} \hat{\sigma}^{2}_t = \tilde \sigma^2, \quad \mbox{a.s.}, 
		\end{equation*}
		where $\hat{\sigma}^2_t$ is the estimator in Eq \eqref{eq:variance for proposed estimator}.
	\end{theorem}

	\begin{proof}
		We define $\theta_{j}(X^{(j)}_i)$ as follows: for $j = 1, 2, \dots$, and $i=1,2, \ldots, n_j$,
		\begin{equation*}
			\begin{split}
				\theta_{j}(X^{(j)}_i)= K^2(X_i^{(j)})w^2_j(X_i^{(j)}) - E_G\left[K^2(X_i^{(j)})w^2_j(X_i^{(j)})\big|\mathcal{F}_{j - 1}\right].
			\end{split}
		\end{equation*}
		Let $\Theta_j = \sum_{i = 1}^{n_j}\theta_{j}(X^{(j)}_i)$.
		Since $E_G[\theta_{j}(X^{(j)}_i)|\mathcal{F}_{j - 1}] = 0$, we obtain that
		\begin{equation*}
			\begin{split}
				E_G[\Theta_j^2|\mathcal{F}_{j - 1}] &= \sum_{i = 1}^{n_j} E_G[\theta^2_{j}(X^{(j)}_i)|\mathcal{F}_{j - 1}],  \\
				& \le  n_j E_G\left[ K^4(X^{(j)})w_j^4(X^{(j)})| \mathcal{F}_{j - 1}\right].
			\end{split}
		\end{equation*}
		Eqs (\ref{ineq:upper for function h}) and (\ref{ineq:upper for weight}) show that
		\begin{equation*}
			E_G\left[ K^4(X^{(j)})w_j^4(X_i^{(j)})| \mathcal{F}_{j - 1}\right] <  \exp \{ 3M_{\rho}|S| \} E\left[ K^4(X)  \prod_{\xi \in X} \frac{\phi^{4}(\xi)}{m_{\rho}^3} \right].
		\end{equation*}
		The right hand side of the above inequality is equal to
		\begin{equation*}
			C =\exp\left \{ (3M_{\rho}-1)|S| + \frac{1}{m_{\rho}^3}\int_S \phi^4(\xi)d\xi\right \} E_{q_3}[K^4(X)],
		\end{equation*}
		which is finite from the assumption of the theorem.
		Since $E_G[E_G[\Theta_{j}^2|\mathcal{F}_{j - 1}] ] = E_G\left[\Theta_j^2\right]$,  we have that
		\begin{equation*}
			E_G\left[\Theta_j^2\right]  <  n_j C.
		\end{equation*}
		By applying Eq (\ref{ineq:upper for sample size}) to the above equation, we obtain that $\sum_{j = 1}^{\infty} E_G\left[\Theta_j^2\right]/(n^{(j)})^2$ is finite. Then,  it follows from Feller~\citeyearpar[Theorem~3, p.243]{feller1introduction} that
		\begin{equation*}
			\lim_{t\rightarrow \infty }\frac{1}{n^{(t)}} \sum_{j = 1}^t \Theta_j = 0.
		\end{equation*}

		We obtain from the above equation that
		\begin{equation*}
			\begin{split}
				&\lim_{t \rightarrow \infty} \frac{1}{n^{(t)}} \sum_{j = 1}^t\sum_{i = 1}^{n_j} K^2(X_i^{(j)})w^2_j(X_i^{(j)})  \\ &=\lim_{t \rightarrow \infty} \frac{1}{n^{(t)}} \sum_{j = 1}^t n_j E_G\left[K^2(X^{(j)})w^2_j(X^{(j)})\big|\mathcal{F}_{j - 1}\right] \quad \mbox{a.s.},
			\end{split}
		\end{equation*}
		if the right hand side of the above equation exists.  We have from Eq~\eqref{eq:convergence of the conditional expectation of  K'(X)w_t(X)}  that
		the right hand side of the above equation converges to $E_{\tilde g}[K^2(X)\tilde w^2(X)]$ almost surely (Etemadi \citeyearpar[Theorem 1]{etemadi2006convergence}). Then, we obtain that
		\begin{equation*} \label{eq:convergence for k_square w_square}
			\lim_{t \rightarrow \infty} \frac{1}{n^{(t)}} \sum_{j = 1}^t\sum_{i = 1}^{n_j} K^2(X_i^{(j)})w^2_j(X_i^{(j)}) = E_{\tilde g}[K^2(X)\tilde w^2(X)], \quad \mbox{a.s.}
		\end{equation*}

		In obtaining the above convergence on the weighted average of $\{K^2(X_i^{(j)}), j=1, 2, \ldots, t,  i = 1, 2, \ldots, n_j\}$,  we have assumed that $E_{q_3}[ K^4(X)]$ is finite.  Due to the Liaponov inequality \citep[p.50]{chung2000course}, $E_{q_3}[ K^2(X)]$ is also finite. Thus, we can replace $K^2(x)$ with $K(x)$ in the above equation, which gives that 
		\begin{equation*} \label{eq:convergence for k w_square}
			\lim_{t \rightarrow \infty} \frac{1}{n^{(t)}} \sum_{j = 1}^t\sum_{i = 1}^{n_j} K(X_i^{(j)})w^2_j(X_i^{(j)}) = E_{\tilde g}[K(X)\tilde w^2(X)], \quad \mbox{a.s.}
		\end{equation*}
		Letting $K(x) = 1$ in the above equation gives that
		\begin{equation*}
			\lim_{t \rightarrow \infty} \frac{1}{n^{(t)}} \sum_{j = 1}^t\sum_{i = 1}^{n_j} w^2_j(X_i^{(j)}) = E_{\tilde g}[\tilde w^2(X)], \quad \mbox{a.s.}
		\end{equation*}

		It follows from the above three equations and Theorem~\ref{thm:the almost sure convergence of mu_t} that
		\begin{equation*}
			\lim_{t \rightarrow \infty} \frac{1}{n^{(t)}} \sum_{j = 1}^t \sum_{i = 1}^{n_j} (K(X_i^{(j)}) - \hat \mu_t)^2 w^2_j(X_i^{(j)}) = E_{\tilde g}[(K(X) - \mu)^2\tilde w^2(X)], \quad \mbox{a.s.}
		\end{equation*}
		Eq~\eqref{eq:convergence for w} and the above equation imply that
		\begin{equation*}
			\lim_{t \rightarrow \infty} \hat{\sigma}_t^2 =  \frac{E_{\tilde{g}}[(K(X) - \mu)^2\tilde w^2(X)]}{E_{\tilde{g}}[\tilde w(X)]^2},  \quad \mbox{a.s.}
		\end{equation*}
		The right hand side of the above equation is equal to $\tilde \sigma^2$ given in Eq~\eqref{eq:the limit of sigma square}.
	\end{proof}
	
	\subsection{Stopping criteria}
	\label{subsec:stopping criteria}
	Oh and Berger \citeyearpar{oh1992adaptive} suggested that the iterative estimation of $\mu$ should continue until  the following condition is satisfied: for a sufficiently small values of $\epsilon$ and $\alpha$,
	\begin{equation*}
		\text{Pr} \left\{\left | \frac{\hat{\mu}_t - \mu}{\mu}  \right | \le \epsilon \right\} \ge 1 - \alpha.
	\end{equation*}
	In other words, if the relative error of $\hat \mu_t$ is very small with a probability close to $1$, then the iterative estimation of $\mu$ stops, and the current value of $\hat \mu_t$ is the estimate to $\mu$.
	Due to the asymptotic normality of $\hat \mu_t$ given in Corollary~\ref{cor:the asymptotic normality of mu_t}, the above condition is converted to  that
	\begin{equation*}
		\frac{\tilde \sigma^2}{n^{(t)}\mu^2} \le \left(\frac{\epsilon}{z_{\alpha/2}}\right)^2.
	\end{equation*}
	By substituting $\hat \mu_t$ for $\mu$ and substituting $\hat{\sigma}_t^2$ for $\tilde \sigma^2$ in the above condition, we obtain the stopping criterion on $\hat \mu_t$ given in Algorithm~\ref{alg:adaptive importance sampling}. In the numerical study in Section~\ref{sec:numerical results}, we set $\epsilon = 0.09$ and $\alpha = 0.01$, which implies that $\eta_1 \approx (0.05)^2$.

	The stopping criterion  obtained above is valid only when $\mu_t$ and  $\hat \sigma^2_t$ are sufficiently close to $\mu$ and $\tilde \sigma^2$, respectively. These events may occur when  $\hat{\rho}_t$ is very close to $\tilde \rho$. In order to check this, we add the following condition on $\hat \rho_t$ to the stopping criterion: for a small $\eta_2 > 0$,
	\begin{equation} \label{ineq:stopping criterion for rho}
		\frac{|\hat{\rho}_t - \hat{\rho}_{t - 1}|}{\hat{\rho}_{t - 1}} \le \eta_2.
	\end{equation}
	If both the stopping conditions described above are satisfied, then  we stop the iterative estimation  of $\mu$, and return the final value of $\hat \mu_t$ as  the estimate to $\mu$.

	\section{Estimation of the intensity of a stationary pairwise interaction point process}
	\label{sec:the intensity of a stationary pairwise interaction point process}

	In this section, we assume that $X$ is a stationary pairwise interaction point process defined on $\mathbb R^d$.  Let $\nu$ be the interaction function of $X$, and $\beta$ be the activity parameter of it. We assume that $\nu(\xi, \eta)$, $\xi, \eta \in \mathbb R^d$, only depends on the distance between $\xi$ and $\eta$, i.e. $\nu(\xi, \eta) = \nu(\parallel \xi -\eta \parallel)$. We also assume that  $X$ is repulsive with finite range of interaction $r$.  Since the number of points of $X$ is infinite on $\mathbb R^d$, the pdf of $X$ in $\mathbb{R}^d$ is not well defined \citep[Chapter 5.5.3]{chiu2013stochastic}.  
	Instead, the Papangelou conditional intensity defined for a finite point process can be extended to $X$ as follows \citep{moller2003statistical}:
	\begin{equation}
		\label{eq:the Papangelou conditional intensity  of stationary pairwise interaction point process}
		\lambda^*(x, \xi) = \beta\prod_{\eta \in x} \nu \left(\parallel \eta - \xi \parallel\right), \quad x \in \mathcal{N}_{\text{lf}}, \; \xi \in \mathbb R^d,
	\end{equation}
	where $\mathcal{N}_{\text{lf}}$ is the set of locally finite point patterns on $\mathbb R^d$, i.e.  $\mathcal{N}_{\text{lf}} = \{x \subset \mathbb R^d : n(x \cap B) < \infty, \text{ for all bounded region } B \in  \mathbb R^d \}$. 
	Since the range of interaction is finite, we can see that the value of $\lambda^*(x, \xi)$ is determined by the  product of finite number of interactions and $\beta$.

	The intensity of $X$ is defined as the expected number of points occurring per unit region, i.e. for any bounded region $B \in \mathbb R^d$,
	\begin{equation*}
		\lambda = \frac{\text{the expected number of points of $X$ on $B$}}{|B|}.
	\end{equation*}
	Clearly, the value of $\lambda$ does not depend on the choice of $B$ due to the stationarity of~$X$. Let $o$ be the origin, and $B(o,r)$ be the ball  with center $o$ and radius $r$. Then, it follows from Eq~\eqref{eq:the Papangelou conditional intensity  of stationary pairwise interaction point process} that  $\lambda^*(X, o)$ is represented as 
	\begin{equation*}
		\lambda^*(X, o) = \beta \prod_{\eta \in X \cap B(o,r) } \nu (\parallel \eta \parallel ).
	\end{equation*}
	Let $f^*$ be the pdf of $X$ on $B(o,r)$.  Then, $\lambda$ is  represented by Georggii-Nguyen-Zessin formula \citep{nguyen1979ergodic, georgii1976canonical} as follows:
	\begin{equation}
		\label{eq:GNZ formula}
		\lambda = E_{f^*}[\lambda^*(X, o)].
	\end{equation}

	Bounded regions  are considered as the domain of the process for the estimation of the summary characteristics such as the intensity and the higher order moments of $X$~\citep{illian2008statistical}). Suppose that $X_S = X \cap S$ is a finite pairwise interaction point process defined on a bounded region $S$, and that $X_S$ has the same parameters as $X$. Then, the pdf of $X_S$ with respect to Poi$(S, 1)$ is defined, for $\beta > 0$, as
	\begin{equation}
		\label{eq:def of the stationary pairwise interaction point process with parameters nu and beta}
		f(x) \propto \beta^{n(x)} \prod_{\{\xi, \eta\} \subseteq x} \nu(\parallel \xi - \eta \parallel), \quad x \in \mathcal{N}.
	\end{equation}
	Then, the Papangelou conditional intensity of $f$ is obtained as
	\begin{equation*}
		\lambda_f(x, \xi) = \beta\prod_{\eta \in x} \nu \left(\parallel \eta - \xi \parallel\right), \quad x \in \mathcal{N}, \; \xi \in S\backslash x.
	\end{equation*}
	Since  $X_S$ is repulsive, $\lambda_f$ is bounded by $\beta$, which implies that $f$ is locally stable.

	We choose $S$ such that it contains the origin, and that it is sufficiently large compared to  $B(o,r)$.  In this case, the Papangelou conditional intensity $\lambda(X_S, o)$ at the origin may have the very similar distribution to $\lambda(X, o)$, which enables us to apply Eq~\eqref{eq:GNZ formula} to $X_S$ instead of $X$ for the estimation of $\lambda$. By letting $K(x) = \lambda_f(x, o)$ and $\mu = E_f[\lambda_f(X, o)]$, we can apply the proposed scheme as well as the MH and the CFTP methods to estimate the value of $\lambda$.

	\section{Numerical results}
	\label{sec:numerical results}
	In this section,  we estimate the intensity for a stationary Strauss point process, and estimate the expected number of points in a region for a non-stationary Strauss point process. We apply the estimators described in Sections~2 and~3; the MH estimator in~Eq~\eqref{eq:MH estimator}, the CFTP estimator in Eq~\eqref{eq:CFTP estimator}, and the AIS estimator in Algorithm~\ref{alg:adaptive importance sampling}. The efficiency of the estimators is compared in terms of the simulation time to get the same level of accuracy.  This type of efficiency can be measured by the product of the squared standard error of the estimate and the simulation time to get the  estimate (Glynn and Whitt~\citeyearpar{glynn1992asymptotic},  Sak and H{\"o}rmann~\citeyearpar{sak2012fast}).  We denote it by time-variance.  The smaller the time-variance of an estimator is, the more efficient the estimator is.  For pair comparison of the time-variance performance, we implemented the algorithms of  three estimation methods using R~\citeyearpar{R_package} instead of using the built-in functions provided by the R package spatstat~\citeyearpar{baddeley2005spatstat}.


	\subsection{Estimation of the intensity of a stationary Strauss point process}
	We consider a finite Strauss point process  $X$ on a bounded region $S = [-0.5, 0.5] \times [-0.5, 0.5]$.  The range of interaction of $X$ is~0.1. Let $o$ be the origin $(0,0)$. It follows from  Eq~\eqref{eq:papangelou conditional intensity of strauss point process} that if $X$ does not contain $o$,  the Papangelou conditional intensity of $X$ is given by
	\begin{equation*}
		\lambda_f(x, o) = \beta \gamma^{\sum_{\eta} I(\parallel \eta \parallel < 0.1)}.
	\end{equation*}

	\begin{table}
		\caption{The estimate to $\lambda$, its standard error, and some metrics related to the corresponding simulation}
		\label{table:stationary Strauss PP with fixed relative standard error}
		\centering
		\begin{tabular}{ c | c | c c c c c c}
			\hline
			\multicolumn{8}{l}{$\beta = 50$}  \\
			\hline
			$\gamma$ & method  & $\hat{\lambda}$ & s.e  & simulation  time  & number of samples & time-variance & t-v ratio \\
			\hline
			\multirow{3}{*}{0.2}
			& MH   & 23.6 & 1.2 & 292   & 313   & 408           & 2.1  \\
			& CFTP & 23.6 & 1.2 & 3211  & 309   & 4510          & 23 \\
			& AIS  & 24.8 & 1.2 & 129   & 9.1$\times 10^4$ & \textbf{199}  & 1\\
			\hline
			\multirow{3}{*}{0.4}
			& MH   & 25.6 & 1.3 & 176    & 184  & 286           & 22  \\
			& CFTP & 28.0 & 1.4 & 1558   & 164  & 3034          & 237 \\
			& AIS  & 28.9 & 1.4 & 6.3      & 4000 & \textbf{13}   & 1   \\
			\hline
			\multirow{3}{*}{0.6}
			& MH   & 30.5 & 1.5 & 92    & 87   &209         & 58   \\
			& CFTP & 31.5 & 1.6 & 566   & 71   &1387        & 385  \\
			& AIS  & 30.0 & 1.5 & 1.7     & 1100 &\textbf{3.6}  & 1     \\
			\hline
			\multirow{3}{*}{0.8}
			& MH   & 38.5 & 1.9 & 34  & 21  & 123         & 44  \\
			& CFTP & 37.2 & 1.8 & 193 & 29  & 629         & 225  \\
			& AIS  & 39.3 & 1.4 & 1.4   & 900 & \textbf{2.8}  & 1   \\
			\hline
			\multicolumn{8}{l}{$\beta = 100$}  \\
			\hline
			\multirow{3}{*}{0.2}
			& MH   & 31.1 & 1.6 & 822   & 558     & \textbf{1981}   & 0.01\\
			& CFTP & 35.4 & 1.8 & $4.0 \times 10^4$ & 460     & $1.3 \times 10^5$      & 0.7\\
			& AIS  & 34.0 & 1.7 & $6.2 \times 10^4$ & $5.2 \times 10^6$ & $1.8 \times 10^5$ & 1    \\
			\hline
			\multirow{3}{*}{0.4}
			& MH   & 38.8 & 1.9 & 445   & 284   & 1670           & 5.1  \\
			& CFTP & 39.9 & 2.0 & $1.2 \times 10^4$ & 220   & $4.8 \times 10^4 $ & 147\\
			& AIS  & 42.3 & 2.1 & 73    & $3.3 \times 10^4$ & \textbf{328}   & 1  \\
			\hline
			\multirow{3}{*}{0.6}
			& MH   & 47.8 & 2.4 & 206  & 125  & 1172           & 25  \\
			& CFTP & 47.2 & 2.3 & 3141 & 82   & $1.7 \times 10^4$     & 368 \\
			& AIS  & 45.9 & 2.3 & 9.0    & 3500 & \textbf{48}    & 1   \\
			\hline
			\multirow{3}{*}{0.8}
			& MH   & 62.6 & 3.1 & 85   & 40   & 826            & 33 \\
			& CFTP & 64.4 & 3.2 & 999  & 34   & $1.0 \times 10^5$  & 410\\
			& AIS  & 62.5 & 2.8 & 3.2    & 1200 & \textbf{25}    & 1  \\
			\hline
		\end{tabular}
	\end{table}
	
	By applying the three estimation methods with $K(x) = \lambda_f(x,o)$,  we have estimated the value of $E_f[\lambda_f(X,o)]$ with various values of $\beta$ and $\gamma$.  Then, these estimated values can be considered as the estimates to $\lambda$, the intensity of the stationary Strauss point process with the same parameters.  The number of sample point processes generated for obtaining an estimate to $\lambda$ was determined so that  the estimate has a relative standard error of about  $0.05$ or less, i.e. s.e./$\hat \lambda \leq 0.05$.  Table~\ref{table:stationary Strauss PP with fixed relative standard error} shows $\hat \lambda_{\text{MH}}$, $\hat \lambda_{\text{CFTP}}$, and $\hat \lambda_{\text{AIS}}$ for each pair of $\beta$ and $\gamma$, and shows their standard errors.  The simulation times  in second and the numbers of samples generated to obtain $\hat{\lambda}_{\text{MH}}$,  $\hat{\lambda}_{\text{CFTP}}$, and  $\hat{\lambda}_{\text{AIS}}$, respectively, are also shown in Table~\ref{table:stationary Strauss PP with fixed relative standard error}.

	In order to get $\hat \lambda_{\text{MH}}$ for a pair of $\beta$ and $\gamma$, samples of $X$ were generated by applying the MH algorithm described in Section~\ref{sec:MC methods}.  We set the probability of a birth proposal to be $0.5$. The generated samples of $X$ follow  the pdf $f$ in Eq~\eqref{eq:strauss point process} asymptotically.  We set the burn-in period to be $3000$ and choose each $200$-th sample of the remaining samples for obtaining $\hat{\lambda}_{\text{MH}}$.  Let $\{Y_1, Y_2, \ldots, Y_n\}$ be the resulting samples. Then, the sample mean of $\{\lambda(Y_1, o), \lambda(Y_2, o), \ldots, \lambda(Y_n, o)\}$ is the MH estimate to $\lambda$.  In order to get $\hat \lambda_{\text{CFTP}}$  for a pair of $\beta$ and $\gamma$, we have generated independent random copies of  $X$ following the pdf $f$ in Eq~\eqref{eq:strauss point process}. In the generation of a sample point process of $X$, we applied the CFTP method.  We computed the value of $\lambda(X, o)$ for each sample of $X$. The sample mean of $\lambda(X, o)$'s is the CFTP estimate to $\lambda$.   In obtaining  $\hat{\lambda}_{\text{AIS}}$, we have generated samples of homogeneous Poisson point processes with varying values of intensity using  Algorithm~\ref{alg:adaptive importance sampling}.  The number of samples in step $t$, $t\geq 1$, was set to be $n_1 = 500$ in the first step, and $n_2 = n_3 =  \ldots = 100$ in the following steps. By setting $\eta_1 = 0.05^2$, the relative standard error of $\hat \lambda_{\text{AIS}}$ was intended to be $0.05$ or less.  We also set $\hat{\rho}_0 = \beta/3$, $m_{\rho} = 10^{-10}$, $M_{\rho} = 10^{10}$, and $\eta_2 = 0.01$.  In the cases of $(\beta, \gamma )= (50,0.8)$ and $(\beta, \gamma )= (100,0.8)$, the relative standard errors of $\hat \lambda_{\text{AIS}}$ are  less than 0.05, the required relative standard error.  In the two cases, more iterations were required to satisfy the other stopping criterion that $|\hat \rho_t - \hat \rho_{t-1}|/\hat \rho_{t-1}$ should be less than $\eta_2$.

	Table~\ref{table:stationary Strauss PP with fixed relative standard error}  shows that the three estimators give similar estimates to $\lambda$ for all pairs of $\beta$ and $\gamma$. Note that both $\hat \lambda_{\text{MH}}$ and  $\hat \lambda_{\text{AIS}}$ are asymptotically unbiased estimator of $\lambda$, while $\hat \lambda_{\text{CFTP}}$ is an unbiased estimator of $\lambda$ for any finite number of samples.  Since the relative standard error of   $\hat \lambda_{\text{CFTP}}$ is  sufficiently small in this example,  the estimated value of $\lambda$ by the CFTP method may be close to the true value of $\lambda$ for given pair of $\beta$ and $\gamma$,  which means that  both $\hat \lambda_{\text{MH}}$ and  $\hat \lambda_{\text{AIS}}$  gave reliable estimates of $\lambda$.
	Thus, we can see that we have done a sufficient number of iterations in obtaining $\hat \lambda_{\text{MH}}$, and that the estimator $\hat \lambda_{\text{MH}}$ worked well. This conclusion  also holds for the estimator $\hat \lambda_{\text{AIS}}$.

	Contrary to the fact that both the CFTP and the MH methods require approximately the same number of samples to get an estimate to $\lambda$, the simulation time to get $\hat \lambda_{\text{CFTP}}$ is much larger than that of $\hat \lambda_{\text{MH}}$. This is  because it takes  a very long time to generate a sample point process of $X$ using the CFTP method compared to the MH method. We can see  from Table~\ref{table:stationary Strauss PP with fixed relative standard error}  that it took  a little time to get an estimate of $\lambda$ using the AIS method except the cases of $\gamma = 0.2$.  Since it takes a very little time to generate a homogeneous Poisson point process, the AIS method is much faster than the MH and the CFTP methods in generation of a sample point process.  This explains why the simulation time of the AIS method is smaller than the other methods except the cases of $\gamma = 0.2$.  In the cases of $\gamma = 0.2$, the required number of samples for $\hat  \lambda_{\text{AIS}}$ to have the desired relative standard error  is huge and it took very long time to generate the required number of samples.

	We can see from Table~\ref{table:stationary Strauss PP with fixed relative standard error}  that the number of samples required for $\hat{\lambda}_{\text{AIS}}$ is larger than those of $\hat \lambda_{\text{MH}}$ and $\hat \lambda_{\text{CFTP}}$ for all pairs of $\beta$ and $\gamma$. This is due to the fact that in the AIS  method  the generated point processes do not  follow the pdf $f$ in Eq \eqref{eq:strauss point process} but follow a homogeneous Poisson point process. This makes the variance of the likelihood ratio $w_t(X)$ for a generated sample $X$ in Algorithm~\ref{alg:adaptive importance sampling} be large, which results in the large value of $\hat \sigma^2_t$ in Algorithm~\ref{alg:adaptive importance sampling}.  By generating a large number of sample point processes,  the AIS method  reduces the relative standard error of $\hat \lambda_{\text{AIS}}$. The sample point processes generated by the CFTP method follow  the pdf $f$  exactly, and those of the MH method follow the pdf $f$  asymptotically. This explains why the two methods require approximately the same number of samples to get the same relative standard error.
	
	\begin{figure}[h]
		\centering
		\begin{subfigure}[b]{0.4\textwidth}
			\centering
			\includegraphics[width=\textwidth]{beta100_gamma02_R01.png}
			\caption{}
		\end{subfigure} \quad
		\begin{subfigure}[b]{0.4\textwidth}
			\centering
			\includegraphics[width=\textwidth]{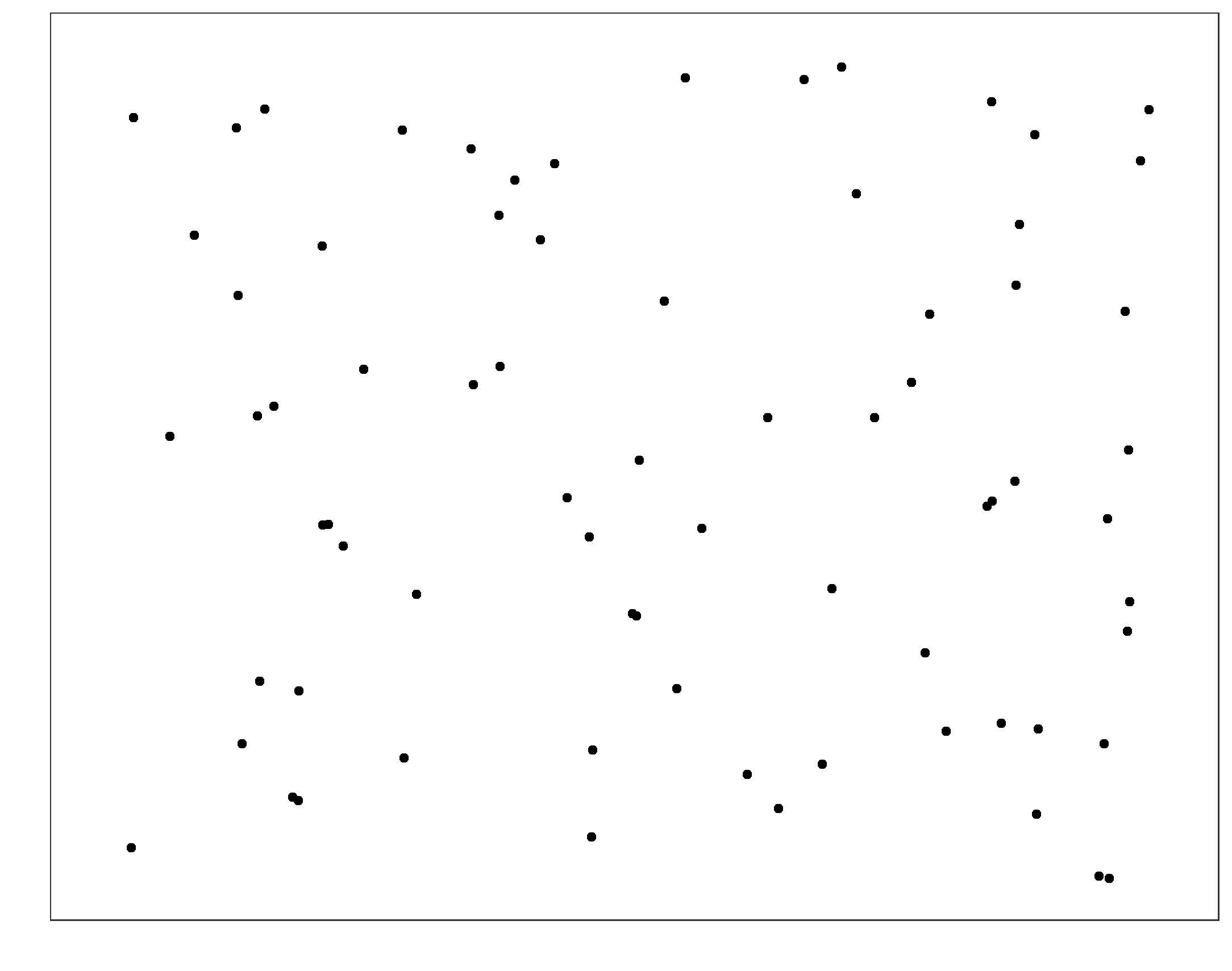}
			\caption{}
		\end{subfigure} \\
		\begin{subfigure}[b]{0.4\textwidth}
			\centering
			\includegraphics[width=\textwidth]{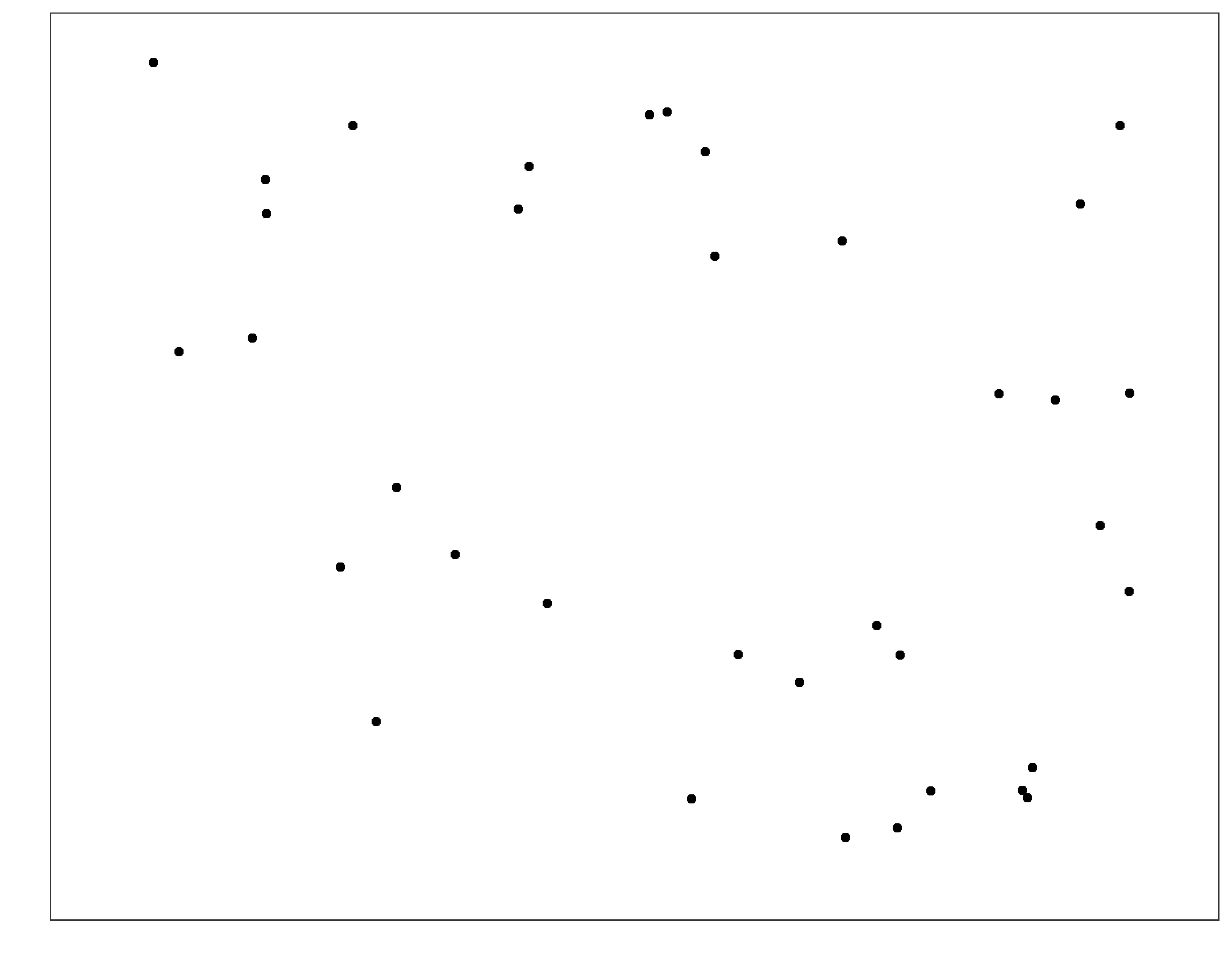}
			\caption{}
		\end{subfigure}\quad
		\begin{subfigure}[b]{0.4\textwidth}
			\centering
			\includegraphics[width=\textwidth]{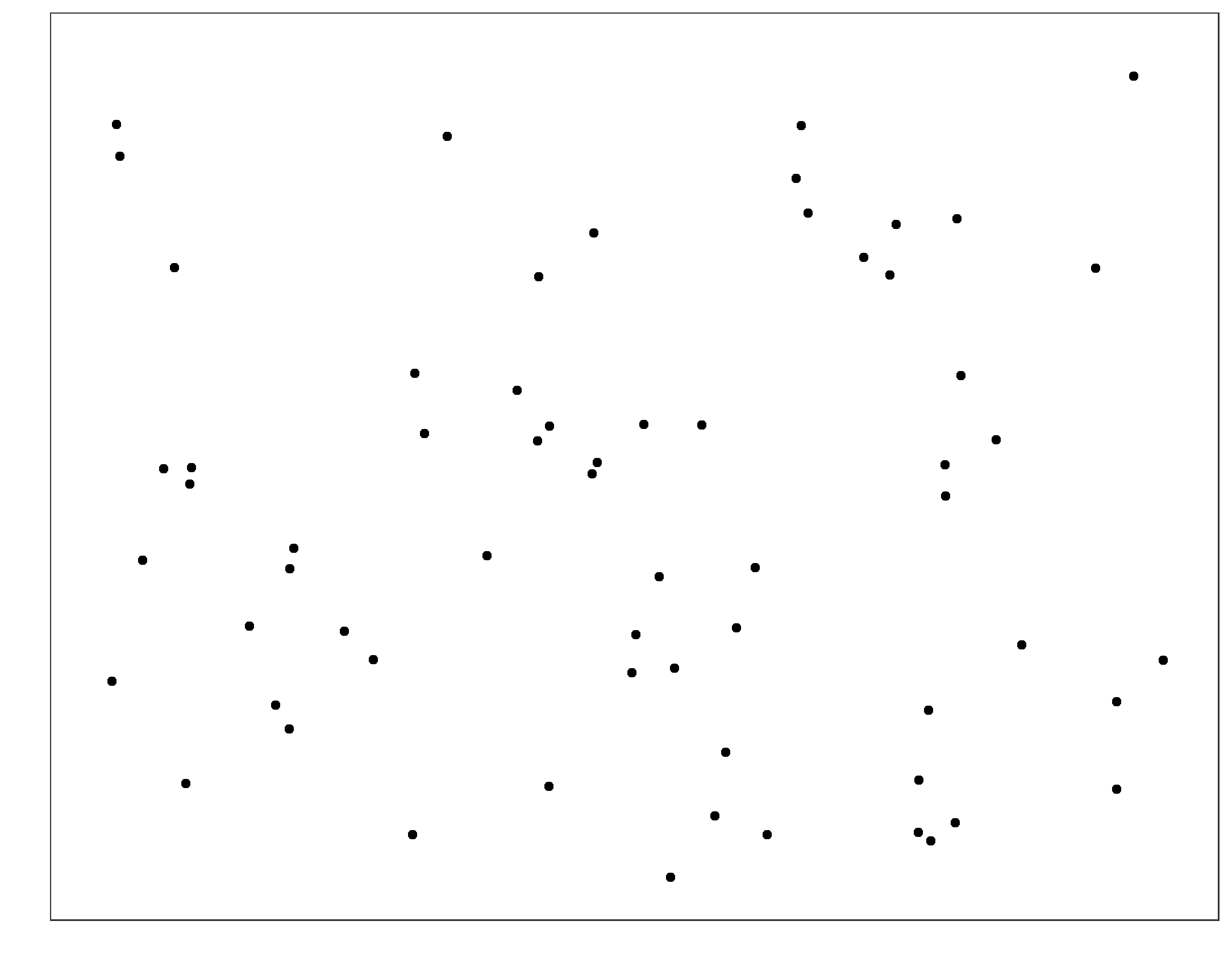}
			\caption{}
		\end{subfigure}
		\caption{Plots of realizations for point processes on the unit square: (a) the stationary Strauss point process with $\beta = 100$ and $\gamma = 0.2$, (b) the stationary Strauss point process with $\beta = 100$ and $\gamma = 0.8$, (c) the Poisson point process with $\rho = 34$, and (d) the Poisson point process with $\rho = 62$. }
		\label{fig:realization for strauss point processes and poisson point process}
	\end{figure}

	Figure~\ref{fig:realization for strauss point processes and poisson point process} illustrates realizations of Strauss point processes  and homogeneous Poisson point processes. The top-left panel of the figure shows a realization of the Strauss point process with $\beta = 100$ and $\gamma = 0.2$, and the  top-right panel  with $\beta = 100$ and $\gamma = 0.8$. The bottom-left panel of the figure  shows a realization of the homogeneous Poisson point process with intensity $34$, which is the estimated value of $\lambda$ of  the Strauss point process shown on  the top-left panel. This is the same with the bottom-right panel, where the intensity is $62$. The estimated $\lambda$'s of the Strauss point processes on  the top panels  can be found  in Table~\ref{table:stationary Strauss PP with fixed relative standard error}.

	We can see that  that the left two panels of 
	Figure~\ref{fig:realization for strauss point processes and poisson point process} look very different from each other. It implies that for a point process $X$ generated from Poi$(S,34)$, the likelihood ratio $w(X)$ computed in Algorithm~\ref{alg:adaptive importance sampling} for the estimation of $\lambda$ of  the Strauss point process shown on the top-left panel may be very small, which results in the degeneracy, and leads a large variance of $\hat \mu_t$. The right two panels  of the figure look similar to each other. It implies that the degeneracy does not occur, and that the variance of the estimator will not take a large value. Thus, we needed much more importance samples of $X$ in the former case in order to get the desired relative error.

	Table~\ref{table:stationary Strauss PP with fixed relative standard error} shows the time-variances of $\hat{\lambda}_{\text{MH}}$, $\hat{\lambda}_{\text{CFTP}}$, and $\hat{\lambda}_{\text{AIS}}$.
	For the cases with $(\beta, \gamma) \neq (100,0.2)$, the table shows that the time-variances of $\hat{\lambda}_{\text{AIS}}$  are   $2$ to $58$ times smaller than those of $\hat{\lambda}_{\text{MH}}$, and they are  $23$ to $410$ times smaller than those of   $\hat{\lambda}_{\text{CFTP}}$.  This means that the AIS method is  $2$ to $58$ times faster than the MH method, and  that it  is  $23$ to $410$ times faster than the CFTP  method in terms of the simulation time to get the same level of accuracy.  Thus, in the case of $(\beta, \gamma) \neq (100,0.2)$,  the AIS method is more efficient than the CFTP and the MH methods.

	For the pair of $(\beta, \gamma) = (100,0.2)$, the ratio of the time-variance   of $\hat{\lambda}_{\text{MH}}$ to that of $\hat{\lambda}_{\text{AIS}}$  is $0.01$. Thus, the MH method is more efficient  than the AIS method in these cases.  The ratio of the time-variance   of $\hat{\lambda}_{\text{CFTP}}$ to that of $\hat{\lambda}_{\text{AIS}}$   for the pair of $(\beta, \gamma) = (100,0.2)$ is $0.7$, which means that the CFTP method is more efficient  than the AIS method in this pair.  In conclusion, we can see that the AIS method is more efficient than the other methods  in the case with large values of $\gamma$,  and that the larger the value of $\gamma$ is, the more efficient the AIS method is.

	\subsection{Estimation of the expected number of points for an inhomogeneous Strauss point process}
	
	In this subsection, we consider an inhomogeneous Strauss point process  $X$  on a bounded region $S = [-0.5, 0.5] \times [-0.5, 0.5]$.  The range of interaction of $X$ is $0.1$. The pdf of $X$ is as follows: for $\beta > 0$, $0 < \gamma < 1$,
	\begin{equation} \label{def:non stationary struass point process}
		f(x; \beta, \gamma) \propto \beta^{n(x)} \gamma^{D(x)} \prod_{\xi \in x} \exp\left( - \alpha \xi_2^2 \right), \quad x \in \mathcal{N},
	\end{equation}
	where $D(x) = \sum_{\{\xi, \eta\}\subset x} I(\parallel \xi - \eta \parallel \le 0.1)$ and $\xi_2$ is the $y$-coordinate of a point $\xi$. Then, the Papangelou conditional intensity is given by
	\begin{equation*}
		\lambda_f (x, \xi) = \beta \gamma^{\sum_{\eta \in x} I\left(\parallel \xi - \eta \parallel \leq 0.1 \right)} \exp\left(-  \xi_2^2\right), \quad x \in \mathcal{N}, \; \xi \in S \backslash x.
	\end{equation*}
	Note that $\lambda_f (x, \xi) \le \beta$. This implies that $X$ is a locally stable point process.
	Let $S_b = \{\xi \in S; |\xi_2| \ge 0.49\}$. If we define $K(x) = \sum_{\xi \in x} I\left( \xi \in S_b \right)$, then the expected number of points of $X$ belonging to $S_b$ is represented as
	\begin{equation*}
		\mu = E_f\left[ K(X) \right].
	\end{equation*}

	\begin{table}
		\caption{The estimate to $\mu$, its standard error, and some metrics related to the corresponding simulation}
		\label{table: estimates to the number of points of the non-stationary Strauss point process}
		\centering
		\begin{tabular}{c | c | c c c c c r}
			\hline
			\multicolumn{8}{l}{$\beta = 50$}  \\
			\hline
			$\gamma$ & method & $\hat{\mu}$ & s.e &  simulation time  & number of samples  &  time-variance & t-v ratio \\
			\hline
			\multirow{3}{*}{0.4}
			& MH   & 0.61 & 0.031 & 861  & 623 & 0.8 & 83 \\
			& CFTP & 0.64 & 0.032 & 6556 & 622 & 6.6 & 678 \\
			& AIS  & 0.61 & 0.030 & 11   & $1.8 \times 10^4$ & \textbf{0.01} & 1 \\
			\hline
			\multirow{3}{*}{0.8}
			& MH   & 0.69 & 0.035  & 1001  & 596 & 1.2 & 891   \\
			& CFTP & 0.72 & 0.036  & 5174  & 579 & 6.7 & 5028  \\
			& AIS  & 0.69 & 0.033  & 1.2     & 2000 & \textbf{0.001} & 1   \\
			\hline
			\multicolumn{8}{l}{$\beta = 100$} \\
			\hline
			\multirow{3}{*}{0.4}
			& MH   & 1.00  & 0.050	 & 790         & 395        & 2.0 & 16          \\
			& CFTP & 1.06  & 0.053  & $6.3 \times 10^5$      & 420    & 1740 &  $1.4 \times 10^4$        \\
			& AIS  & 0.96  & 0.048  & 54          & $8.8 \times 10^4$ & \textbf{0.12} & 1 \\
			\hline
			\multirow{3}{*}{0.8}
			& MH   & 1.16 & 0.058 & 963     & 342  & 3.3 & 771 \\
			& CFTP & 1.46 & 0.073 & $1.4 \times 10^5$     & 261  & 723 & $1.7 \times 10^5$  \\
			& AIS  & 1.20 & 0.059 & 1.2        & 1800 & \textbf{0.004} & 1\\
			\hline
		\end{tabular}
	\end{table}

	For various values of $\beta$ and $\gamma$, we estimated the value of $\mu$ by the three estimation methods.  As in the previous example, the number of sample point processes generated for obtaining an estimate  was determined so that  the estimate has a relative standard error of about  $0.05$ or less.  In obtaining  $\hat{\mu}_{\text{MH}}$, $\hat{\mu}_{\text{CFTP}}$, and  $\hat{\mu}_{\text{AIS}}$,  we applied the same procedures as described in the previous subsection. Table~\ref{table: estimates to the number of points of the non-stationary Strauss point process} shows $\hat \mu_{\text{MH}}$, $\hat \mu_{\text{CFTP}}$, and $\hat \mu_{\text{AIS}}$ for each pair of $\beta$ and $\gamma$, and shows their standard errors.  The simulation times  and the numbers of samples generated to obtain $\hat{\mu}_{\text{MH}}$,  $\hat{\mu}_{\text{CFTP}}$, and  $\hat{\mu}_{\text{AIS}}$, respectively, are also shown in Table~\ref{table: estimates to the number of points of the non-stationary Strauss point process}.

	In Table~\ref{table: estimates to the number of points of the non-stationary Strauss point process},  we can see that the three estimators gave similar estimates to $\mu$ for all pairs of $\beta$ and $\gamma$. In the same reason as in the previous subsection, $\hat \mu_{\text{CFTP}}$ is a reliable estimate to $\mu$, which implies that $\hat{\mu}_{\text{MH}}$ and $\hat{\mu}_{\text{AIS}}$ are also  reliable estimates to $\mu$. As expected, the number of samples for $\hat{\mu}_{\text{AIS}}$ is much  larger than those of $\hat{\mu}_{\text{MH}}$ and $\hat{\mu}_{\text{CFTP}}$ for all pairs of $\beta$, $\gamma$. However,  it took very little time  to get a single estimate to $\mu$ using the AIS method compared to those of  the CFTP and the MH methods for all pairs of $\beta$ and $\gamma$.

	Table~\ref{table: estimates to the number of points of the non-stationary Strauss point process} shows that the time-variances of $\hat{\mu}_{\text{AIS}}$ are 16 to 891 times smaller than those of $\hat{\mu}_{\text{MH}}$, and they are 678 to $1.7 \times 10^5$ times smaller than those of $\hat{\mu}_{\text{CFTP}}$. In order to get an estimate with the same relative standard error, the AIS method is   16 to 891 times faster than the MH method, and that it is  678 to $1.7 \times 10^5$ times faster than the CFTP method in terms of the simulation time. Thus, the AIS method is more efficient than both  the CFTP and the MH methods.

	\section{Conclusion}
	\label{sec:conclusion}
	For a statistic of a locally stable point process in a bounded region, we proposed an adaptive importance sampling for the  estimation of the expected value of  the statistic. In our proposal, the  importance sampling point process is restricted to the family of homogeneous Poisson point processes. We applied the cross-entropy minimization method to find the optimal intensity of the importance point process. In the proposed scheme,  the expected value of the statistic and the optimal intensity are iteratively estimated in an adaptive manner.  We proved the consistency of the proposed estimator and the asymptotic normality of it. In the numerical study, we considered two examples, in which we applied the proposed scheme to the estimation of the intensity of a stationary Strauss point process, and to the estimation of the expected number of points belonging to a subregion for an inhomogeneous Strauss point process on a bounded region.  Numerical results showed that the proposed estimator is more efficient than the CFTP and the MH estimators in terms of the time-variance performance.  This is due to the fact that it takes very little time to generate a sample point process following a homogeneous Poisson point process.
	However, if the stochastic property of the nominal density is too different from that of the homogeneous Poisson point process, then the variance of likelihood ratio becomes large, which results in requiring  a large number of samples in order to get a reliable estimate, and thus makes the proposed scheme inefficient.

	As for further research, we suggest to apply the sequential importance resampling when generating sample point processes. Through the method, we expect to generate sample point processes whose stochastic property is similar to that of the nominal point process, or to  that of a desired distribution.
	
	\bmhead{Acknowledgements}
	This research was supported by the 2023 Research Fund of the University of Seoul.
	We would like to thank  the three anonymous referees. Their comments on the first draft of the paper are critical and helpful for the great improvement in the representation of the paper. 
	\bibliography{adaptive_is.bib}
	
\end{document}